\documentclass[letterpaper, 10 pt, conference]{ieeeconf}  

\IEEEoverridecommandlockouts

\usepackage{graphics} 
\usepackage{graphicx}
\usepackage{subfigure}

\usepackage{epsfig} 
\usepackage{amssymb}  

\usepackage{booktabs}
\usepackage{multirow}
\usepackage{pifont}
\usepackage{cite}

\usepackage{latexsym}
\usepackage{bbm}
\usepackage{amsmath, amssymb}
\usepackage{epsfig}

\def\statespace {{\cal S}}
\def\mdp {{\cal M}}
\def\actionspace {{\cal A}}
\def\probs {{\cal P}}
\def\transitionmodel {{\cal T}}

\usepackage{tikz}
\usepackage{textcomp}
\usepackage{hyperref}
\usepackage{lipsum}

\newcommand\copyrighttext{%
  \footnotesize \textcopyright\, 2020 IEEE. Personal use of this material is permitted.  Permission from IEEE must be obtained for all other uses, in any current or future media, including reprinting/republishing this material for advertising or promotional purposes, creating new collective works, for resale or redistribution to servers or lists, or reuse of any copyrighted component of this work in other works.} 
  
\newcommand\copyrightnotice{%
\begin{tikzpicture}[remember picture,overlay]
\node[anchor=south,yshift=10pt] at (current page.south) {\fbox{\parbox{\dimexpr\textwidth-\fboxsep-\fboxrule\relax}{\copyrighttext}}};
\end{tikzpicture}}

\title{\LARGE{\bf MVP: Unified Motion and Visual Self-Supervised Learning for Large-Scale Robotic Navigation}}

\author{Marvin Chanc\'an$^{1,2}$ and Michael Milford$^{1}$
\thanks{The work of M.C. was supported by the Peruvian Government. The work of M.M. was supported by ARC grants FT140101229, CE140100016, the QUT Centre for Robotics, and the Australian Government via grant AUSMURIB000001 associated with ONR MURI grant N00014-19-1-2571.}
\thanks{$^{1}$ QUT Centre for Robotics, School of Electrical Engineering and Robotics, Queensland University of Technology, Brisbane, Australia}
\thanks{$^{2}$ School of Mechatronics Engineering, Universidad Nacional de Ingenier\'ia, Lima, Peru. {\tt\small mchancanl@uni.pe}}}

\begin{document}

\maketitle
\copyrightnotice
\thispagestyle{empty}
\pagestyle{empty}

\begin{abstract}
Autonomous navigation emerges from both motion and local visual perception in real-world environments. However, most successful robotic motion estimation methods (e.g. VO, SLAM, SfM) and vision systems (e.g. CNN, visual place recognition--VPR) are often separately used for mapping and localization tasks. Conversely, recent reinforcement learning (RL) based methods for visual navigation rely on the quality of GPS data reception, which may not be reliable when directly using it as ground truth across multiple, month-spaced traversals in large environments. In this paper, we propose a novel motion and visual perception approach, dubbed \textit{MVP}, that unifies these two sensor modalities for large-scale, target-driven navigation tasks. Our MVP-based method can learn faster, and is more accurate and robust to both extreme environmental changes and poor GPS data than corresponding vision-only navigation methods. MVP temporally incorporates compact image representations, obtained using VPR, with optimized motion estimation data, including but not limited to those from VO or optimized radar odometry (RO), to efficiently learn self-supervised navigation policies via RL. We evaluate our method on two large real-world datasets, Oxford Robotcar and Nordland Railway, over a range of weather (e.g. overcast, night, snow, sun, rain, clouds) and seasonal (e.g. winter, spring, fall, summer) conditions using the new \textit{CityLearn} framework; an interactive environment for efficiently training navigation agents. Our experimental results, on traversals of the Oxford RobotCar dataset with no GPS data, show that MVP can achieve 53\% and 93\% navigation success rate using VO and RO, respectively, compared to 7\% for a vision-only method. We additionally report a trade-off between the RL success rate and the motion estimation precision, suggesting that vision-only navigation systems can benefit from using precise motion estimation techniques to improve their overall performance.
\end{abstract}

\section{Introduction}\label{sec:intro}

Navigation is a key component for enabling the deployment of mobile robots and autonomous vehicles in real-world environments. Current large-scale, real-world navigation systems rely on the usage of GPS data only as ground truth for sensory image labeling \cite{IEEEexample:mirowski2018learning,IEEEexample:ma2019towards,IEEEexample:streetinst,IEEEexample:hermann2019learning,IEEEexample:Chen_2019_CVPR, IEEEexample:talkwalk}. They then reduce the problem of navigation to vision-only methods \cite{IEEEexample:mirowski2018learning}, GPS-level localization combined with publicly available maps \cite{IEEEexample:ma2019towards}, or extend it with language-based tasks \cite{IEEEexample:streetinst,IEEEexample:hermann2019learning,IEEEexample:Chen_2019_CVPR,IEEEexample:talkwalk}. These end-to-end learning approaches are hard to train due to their large network models and weakly-related input sensor modalities. Moreover, their generalization capabilities to environments with different visual conditions is not well explored. In contrast, we have recently shown an alternative non end-to-end vision-based approach using preprocessed compact image representations to achieve practical training and deployment on real data with challenging environmental transitions \cite{IEEEexample:chancan2020citylearn}.

\begin{figure}[!t]
   \vspace{2mm}
   \centering
   \includegraphics[width=\columnwidth]{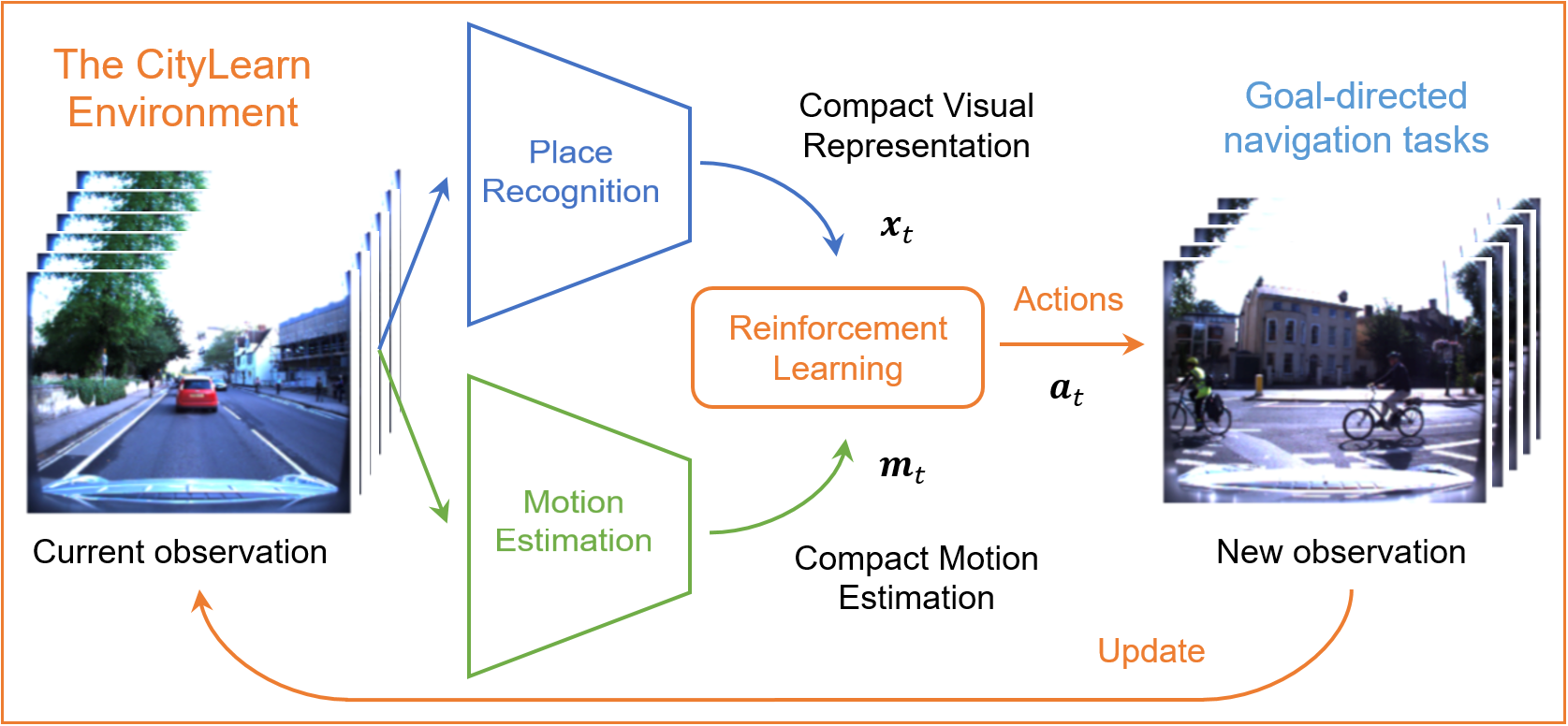}
   \vspace{-7mm}
   \caption{\textbf{Proposed MVP-based approach}. We temporally incorporate odometry-based motion estimation data with compact image representations to perform large-scale all-weather navigation via reinforcement learning. Our method is efficient, accurate and robust to extreme environmental changes, even when GPS data reception fail (see Fig. \ref{influence}).}
   \label{approach}
   \vspace{-4mm}
\end{figure}

\begin{figure}[!t]
   \centering
   \subfigure{\includegraphics[width=0.63\columnwidth]{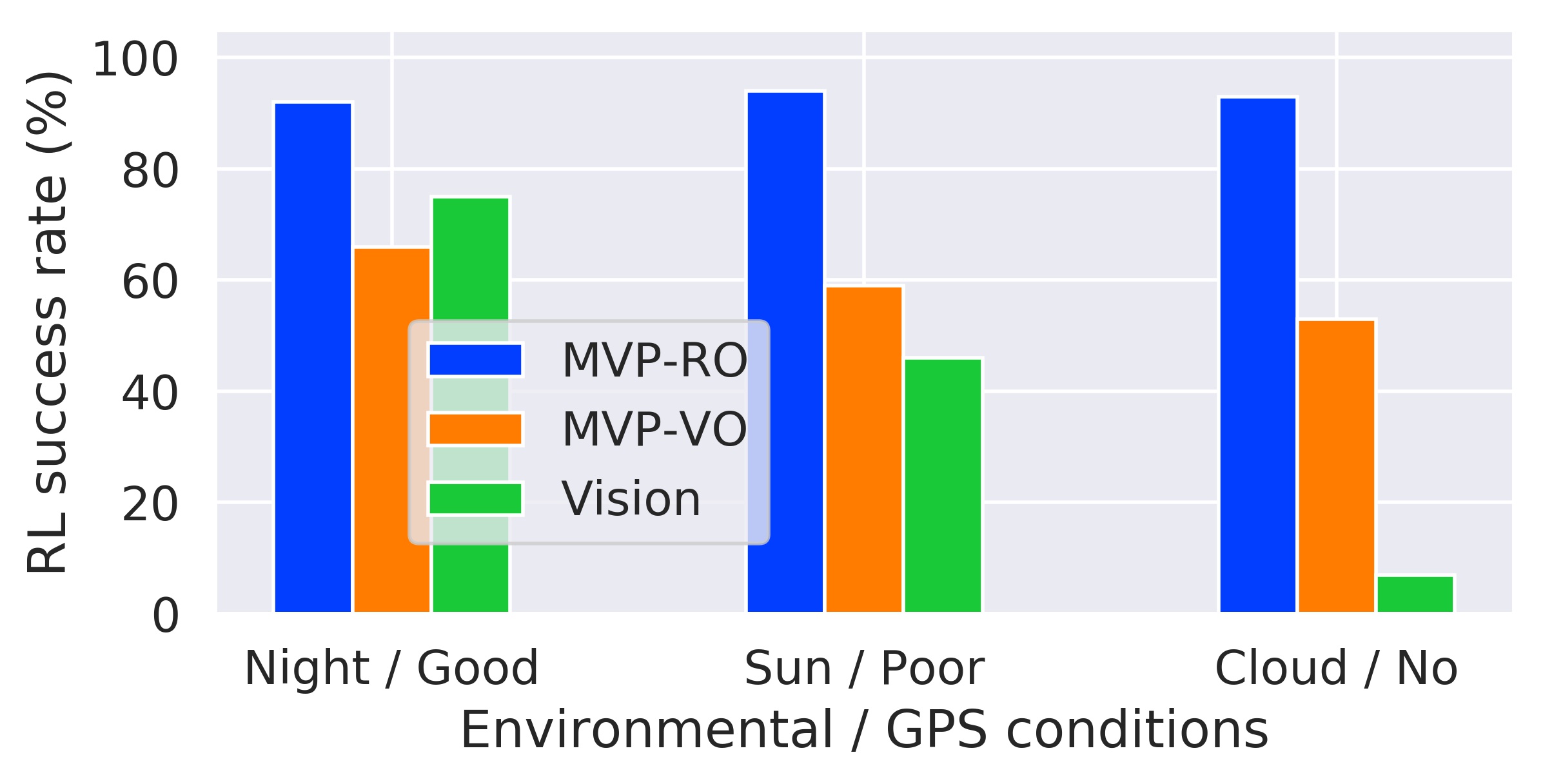}}
   \subfigure{\includegraphics[width=0.33\columnwidth]{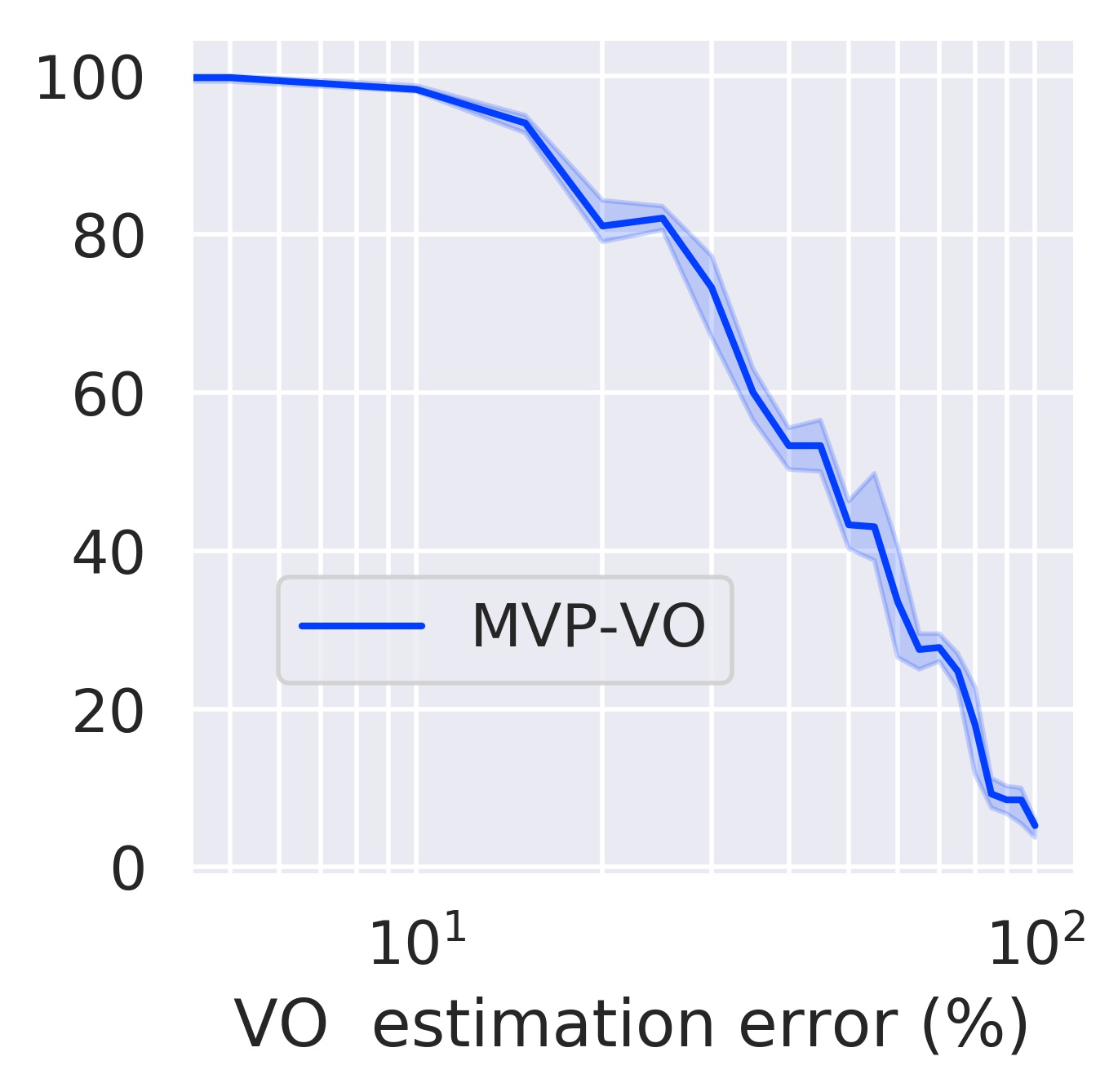}}
   \vspace{-3mm}
   \caption{\textbf{Navigation success rate results on the Oxford RobotCar dataset}. \textbf{Left}: Our MVP-based methods can accurately navigate across a range of visual environmental changes and GPS data situations, where vision-only approaches typically fail. \textbf{Right}: Trade-off curve of the RL success rate and the motion estimation precision using VO (log scale for better visualization).}
   \label{influence}
   \vspace{-4mm}
\end{figure}

In this paper, we build on the main ideas of our previous work \cite{IEEEexample:chancan2020citylearn}---that combines reinforcement learning (RL) and visual place recognition (VPR) techniques for navigation tasks---to present a new, more efficient and robust approach. The main contributions of this paper are detailed as follows:

\begin{itemize}
    \item Leveraging successful robotic motion estimation methods including VO \cite{IEEEexample:mohamed2019surveyodometry} or radar \cite{IEEEexample:barnes2019oxford} to capture compact motion information through an environment that can then be used to perform goal-driven navigation tasks (see Fig. \ref{approach}). This makes our system more efficient and robust to extreme environmental changes, even with limited or no GPS data availability (Fig. \ref{influence}-left).
    \item Using RL to temporally incorporate compact motion representations with equally compact image observations, obtained via deep-learning-based VPR models \cite{IEEEexample:netvlad}, for large-scale, all-weather navigation tasks. 
    \item Experimental results on the RL navigation success rate and the VO motion estimation precision trade-off (Fig. \ref{influence}-right). This shows how our proposed navigation system can improve its overall performance based on precise motion estimation techniques such as VO.
\end{itemize}

We evaluate our motion and visual perception (MVP) method using our interactive \textit{CityLearn} framework \cite{IEEEexample:chancan2020citylearn}, and present extensive experiments on two large real-world driving datasets, the Oxford RobotCar \cite{IEEEexample:maddern2017oxford} and the Nordland Railway \cite{IEEEexample:nordland} datasets. The results of our MVP-based approach are consistently high across multiple, month-spaced traversals with extreme environmental changes, such as winter, spring, fall, and summer for Nordland, and overcast, night, snow, sun, rain, and clouds for Oxford (see blue bar in Fig. \ref{influence}). For Nordland, we show how our approach outperforms corresponding vision-only navigation methods under extreme environmental changes, especially when GPS data is fully available and consistent across multiple traversals of the same route. For Oxford, we show the robustness of our approach across a range of real GPS data reception situations, including poor and no data reception at all, where vision-only navigation systems typically fail.

\section{Related Work}\label{sec:related-work}

We present a brief overview of some successful work in motion estimation research, related visual place recognition methods for sequence-based driving datasets, and recent RL-based navigation systems for large real-world environments.

\subsection{Motion Estimation in Robotics}\label{sub:motion}

Odometry-based sensors (i.e. wheel, inertial, laser, radar, and visual) for \textit{self-localization} have long attracted attention in robotics research as an alternative approach to estimate motion information, especially in situations where GPS data is not reliable such as multi-path reception and changes in environmental conditions \cite{IEEEexample:maddern2017oxford,IEEEexample:mohamed2019surveyodometry,IEEEexample:cadena2016slam}. Traditional VO methods \cite{IEEEexample:vo2004cvpr}, SLAM-based systems \cite{IEEEexample:dissanayake2001radar}, including MonoSLAM \cite{IEEEexample:davison2007monoslam} and ORB-SLAM \cite{IEEEexample:orbslam}, and also bio-inspired models such as RatSLAM \cite{IEEEexample:ratslam2004,IEEEexample:ratslam2008} have captured the main challenges of the localization problem in large outdoor environments and indoor spaces---also with a range of alternative systems \cite{IEEEexample:wifi2010,IEEEexample:renato2016,IEEEexample:renato2017,IEEEexample:renato2019ral}. These methods have shown good invariance to changes in viewpoint and illumination by associating hand-crafted visual place features to locally optimized maps. Odometry, however, is known to be the first phase of solving a SLAM problem, which also includes loop closing and global map optimization. Consequently, multi-sensor fusion techniques combining vision \cite{IEEEexample:orbslam2, IEEEexample:muratal2017vimslam,IEEEexample:vo,IEEEexample:doh2006relative,IEEEexample:pascoe2017nid,IEEEexample:kim2018lowdriftvo,IEEEexample:zhan2019visual}, inertial sensors \cite{IEEEexample:svo,IEEEexample:hong2017vio,IEEEexample:babu2018vio}, LiDAR \cite{IEEEexample:zhang2015visionlidar,IEEEexample:borges2018posemap,IEEEexample:wang2019vilaser}, and radar \cite{IEEEexample:cen2018radar} have been proposed to further improve both the localization accuracy and the real-time performance, with a number of recent deep-learning-based methods that can match the precision of those traditional methods \cite{IEEEexample:wang2017learnVO, IEEEexample:pillai2017learnvo,IEEEexample:zhou2017learnvo,IEEEexample:zhan2018unsupervised,IEEEexample:Casser_2019_CVPR_Workshops,IEEEexample:wang2019learnvo,IEEEexample:loo2019cnn-vo,IEEEexample:shen2019learnvo,IEEEexample:Shen_2019_ICCV,IEEEexample:barnes2019oxford, IEEEexample:suaftescu2020kidnapped}. In this work, we demonstrate our approach using both learning- and conventional-based odometry methods for motion estimation when GPS data is not available, as occurs in several traversals of the Oxford RobotCar dataset \cite{IEEEexample:maddern2017oxford}.

\subsection{Visual Place Recognition}\label{sub:vpr-related}

VPR approaches can be broadly split into two categories: image-retrieval-based methods that compare a single-frame (query) to an image database (reference) \cite{IEEEexample:Torii_2015_CVPR,IEEEexample:weyand2016planet,IEEEexample:netvlad,IEEEexample:2017geo,IEEEexample:gordo2017end,IEEEexample:fine2019imret}, and multi-frame-based VPR techniques built on top of those single-frame methods to work on image sequences \cite{IEEEexample:seqslam,IEEEexample:pepperell2014all,IEEEexample:cnnlanmark,IEEEexample:sunderhauf2015performance,IEEEexample:mpf}; typically found in driving datasets \cite{IEEEexample:nordland,IEEEexample:geiger2012kitti,IEEEexample:maddern2017oxford,IEEEexample:guo2018safe,IEEEexample:naseer,IEEEexample:fabrat,IEEEexample:kitti,IEEEexample:caesar2019nuscenes}. These two approaches often first require computing image descriptors using diverse hand-crafted- \cite{IEEEexample:vprsurvey} or deep-learning-based models \cite{IEEEexample:chen2014convolutional,IEEEexample:zetao2017,IEEEexample:lost}. However, using large image representations can be computationally expensive and also limit the deployment of these methods on real robots. Alternatively, we have recently demonstrated how compact image representations can be used to achieve state-of-the-art results in visual localization \cite{IEEEexample:chancan2020hybrid} by modeling temporal relationships between consecutive frames to improve the performance of compact single-frame-based methods. In this paper, we build on these main ideas to propose our MVP-based approach that uses compact but rich image representations, such as those from NetVLAD \cite{IEEEexample:netvlad}, and can also temporally use movement data through an environment via odometry-based techniques.

\subsection{Learning-based Navigation}\label{sub:navigation}

Significant progress has recently been made in goal-driven navigation tasks using learning methods \cite{IEEEexample:mirowski2018learning,IEEEexample:ma2019towards,IEEEexample:streetinst,IEEEexample:hermann2019learning,IEEEexample:Chen_2019_CVPR,IEEEexample:talkwalk,IEEEexample:kahn2018composable, IEEEexample:mirowski2016complex,IEEEexample:chaplot2018active,IEEEexample:zhang2017neural,IEEEexample:khan2018drlnav,IEEEexample:oh2019learnaction,IEEEexample:kahn2020badgr,IEEEexample:gupta2017cognitive,IEEEexample:gupta2017unifying,IEEEexample:chen2019learning,IEEEexample:chaplot2020Learning,IEEEexample:banino2018vector}, inspired by advances in deep RL \cite{IEEEexample:guo2014atari,IEEEexample:mnih2015human}. Most of these algorithms can successfully train navigation agents end-to-end based on raw images. These approaches, however, are typically only evaluated using either synthetic data \cite{IEEEexample:kahn2018composable, IEEEexample:mirowski2016complex,IEEEexample:chaplot2018active,IEEEexample:zhang2017neural,IEEEexample:banino2018vector}, indoor spaces \cite{IEEEexample:khan2018drlnav,IEEEexample:oh2019learnaction} or relatively small outdoor environments \cite{IEEEexample:kahn2020badgr}, that generally do not require GPS data or map information. Alternatively, combining map-like- or SLAM-based input modalities, including motion sensor data, and images for goal-driven navigation tasks has been proposed \cite{IEEEexample:gupta2017cognitive,IEEEexample:gupta2017unifying,IEEEexample:chen2019learning,IEEEexample:chaplot2020Learning,IEEEexample:mirowski2018learning}, but again these methods are trained only using small indoor environments. For large-scale outdoor navigation, however, different approaches that rely on GPS data as ground truth have been proposed \cite{IEEEexample:mirowski2018learning}, with a range of developments using language-based tasks \cite{IEEEexample:streetinst,IEEEexample:hermann2019learning,IEEEexample:Chen_2019_CVPR,IEEEexample:talkwalk} or publicly available maps \cite{IEEEexample:ma2019towards}. However, relying on GPS data only for benchmarking purposes may not be reliable; especially when using large driving datasets recorded over many month-spaced traversals, as highlighted in previous work \cite{IEEEexample:maddern2017oxford, IEEEexample:barnes2019oxford}.

In this paper, we propose a different approach that overcomes the limitations of prior work for large-scale, all-weather navigation tasks. We unify two fundamental and highly-related sensor modalities: motion and visual perception (MVP) information. Our MVP-based method builds on the main ideas presented in our previous works \cite{IEEEexample:chancan2020citylearn,IEEEexample:chancan2020hybrid}---that use compact image representations to achieve sample-efficient RL-based visual navigation using real data \cite{IEEEexample:chancan2020citylearn}, and also demonstrate how to leverage motion information for VPR tasks \cite{IEEEexample:chancan2020hybrid}. We propose a network architecture that can incorporate motion information with visual observations via RL to perform accurate navigation tasks under extreme environmental changes and with limited or no GPS data; where visual-based navigation approaches typically fail. We provide extensive experimental results in both visual place recognition and navigation tasks, using two large real-world dataset, that show how our method efficiently overcomes the limitations of those vision-only navigation pipelines.

\section{MVP-based Method Overview}\label{sec:method}

Our objective is to train an RL agent to perform goal-driven navigation tasks across a range of real-world environmental conditions, especially under poor GPS data conditions. We therefore developed an MVP-based approach that can be trained using motion estimation and visual data gathered in large environments. Our MVP method operates by temporally associating local estimates of motion with compact visual representations to efficiently train our policy network. Using this data, our policy can learn to associate motion representations with visual observations in a self-supervised manner, enabling our system to be robust to both visual changing conditions and poor GPS data.

In the following sections, we describe our problem formulation via RL, the driving datasets we used in our experiments, details of our MVP representations, our evaluation metrics for VPR and navigation tasks, and related visual navigation methods against which we compare our approach.

\subsection{Problem Formulation}\label{sub:formulation}

We formulate our goal-driven navigation tasks as a Markov Decision Process $\mdp$, with discrete state space $\mathbf{s}_t \in \statespace$, discrete action space $\mathbf{a}_t \in \actionspace$, and transition operator $\transitionmodel: \statespace \times \actionspace \to \statespace$ as in a finite-horizon $T$ problem. Our goal is to find $\theta^*$ that maximizes this objective function:

\begin{equation}
\label{objective}
J(\theta) = \mathbb{E}_{\tau \sim \pi_\theta(\tau)} \left[\sum_{t=1}^T \gamma r(\tau)\right]
\end{equation}

where $\pi_\theta: \statespace \to \probs(\actionspace)$ is the stochastic policy we want to learn, and $r: \statespace \times \actionspace \to \mathbb{R}$ is the reward function with discount factor $\gamma$. We parametrize our navigation policy $\pi_\theta$ with a neural network that can learn $\theta$ to optimize our policy. We also defined our state space $\statespace$ by our compact bimodal MVP space representation ($\mathbf{m}_t$, $\mathbf{x}_t$), and our action space $\actionspace$ by discrete action movements in the agent action space ($\mathbf{a}_t$). 

\subsection{Real-World Driving Datasets}\label{sub:datasets}

We evaluate our approach using our interactive CityLearn framework \cite{IEEEexample:chancan2020citylearn} on two challenging large real-world datasets, the Oxford RobotCar dataset \cite{IEEEexample:maddern2017oxford} and the Nordland Railway dataset \cite{IEEEexample:nordland}, that include diverse environmental changes and real GPS data reception situations.

\begin{figure}[!t]
\centering
\includegraphics[width=\columnwidth]{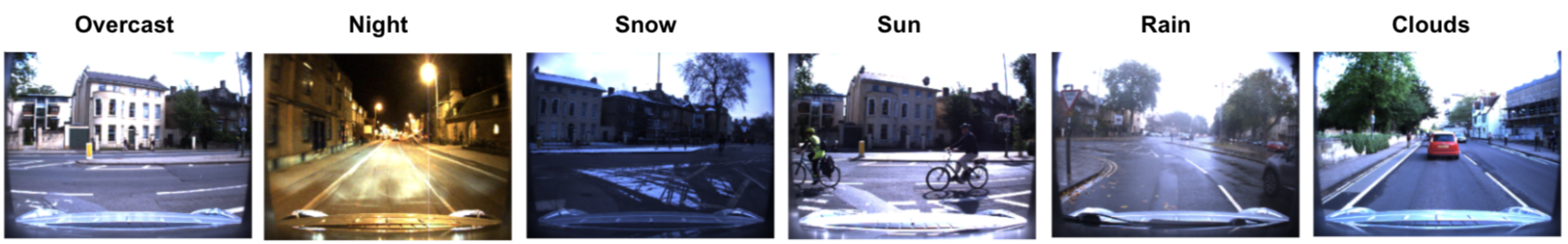}
\vspace{-6mm}
\caption{Our six selected traversals from the Oxford RobotCar dataset.}
\label{oxf-travs}
\vspace{-2mm}
\end{figure}

\begin{figure}[!t]
\centering
\subfigure{\includegraphics[width=0.49\columnwidth]{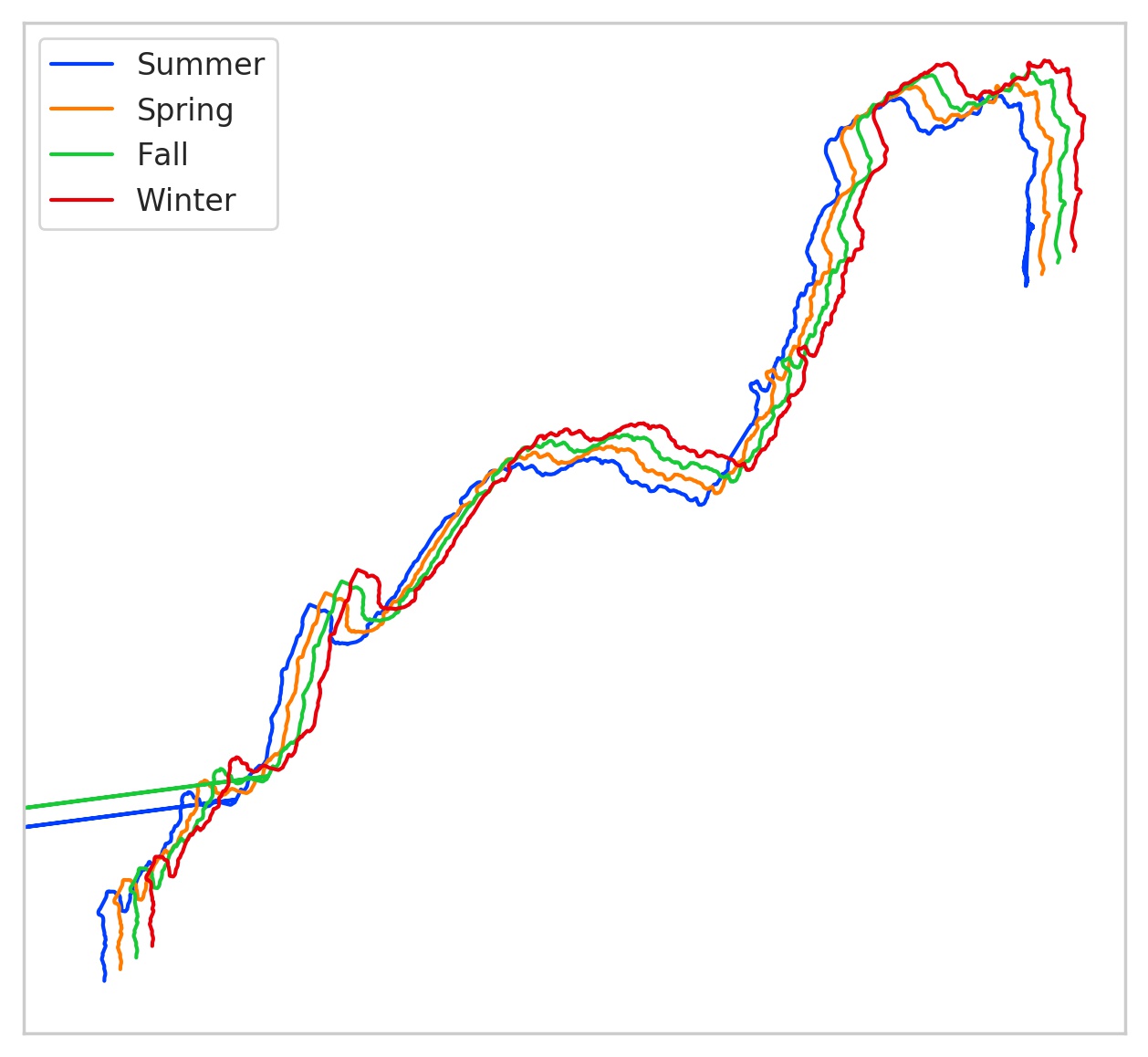}}
\subfigure{\includegraphics[width=0.49\columnwidth]{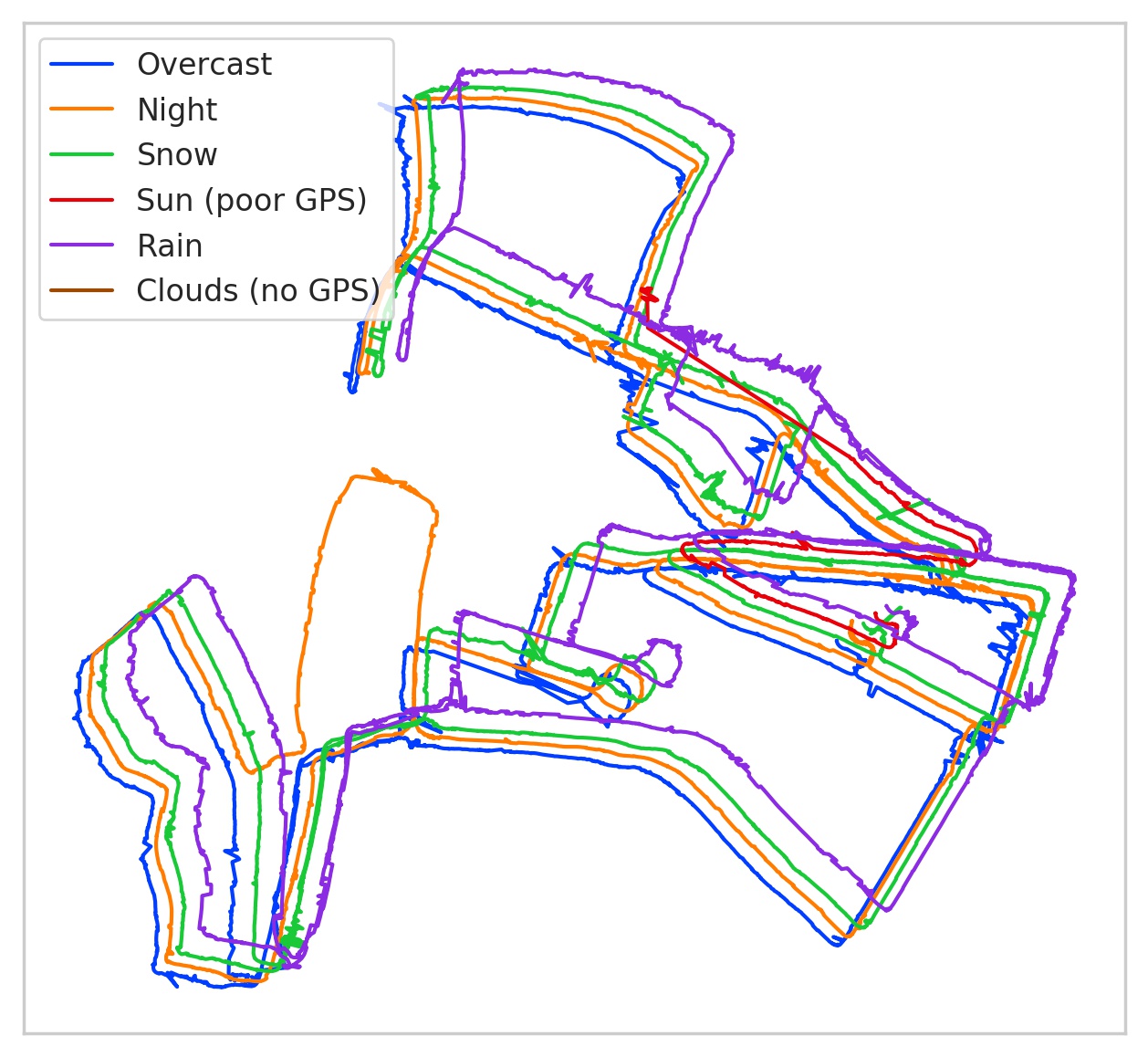}}
\vspace{-6mm}
\caption{\textbf{Raw GPS data}: The 4 traversals from the Nordland Railway (left) dataset and our 6 selected traversals from the Oxford RobotCar (right) dataset, both with a number of weather/season changes and GPS data reception situations (drifted for better visualization).}
\label{gps}
\vspace{-4mm}
\end{figure}

\textbf{Oxford RobotCar}: This dataset \cite{IEEEexample:maddern2017oxford} was collected using the Oxford RobotCar platform over a 10km route in central Oxford, UK. The data recorded with a range of sensors (e.g. LiDARs, monocular cameras and trinocular stereo cameras) includes more than 100 traversals (image sequences) of the same route with a large range of transitions across weather, season and dynamic urban environment changes over a period of 1 year. In Fig. \ref{oxf-travs} we show the selected 6 multiple traversals used in our experiments, referred here as \textit{overcast}, \textit{night}, \textit{snow}, \textit{sun}, \textit{rain}, and \textit{clouds}.\footnote{Referred as 2015-05-19-14-06-38, 2014-12-10-18-10-50, 2015-02-03-08-45-10, 2015-08-04-09-12-27, 2015-10-29-12-18-17, and 2015-05-15-08-00-49, respectively, in \cite{IEEEexample:maddern2017oxford}.} Fig. \ref{gps}-right shows the raw GPS data of our selected traversals, where both the \textit{sun}, and the \textit{clouds} traversals have poor GPS data reception and no GPS data at all, respectively.

\begin{figure}[!t]
\centering
\includegraphics[width=0.96\columnwidth]{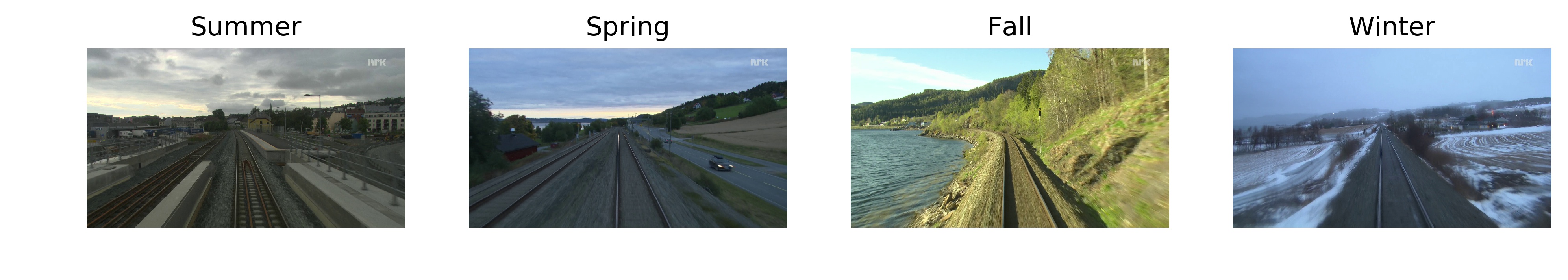}
\vspace{-4mm}
\caption{Samples of the four traversals provided by the Nordland dataset.}
\label{nord-travs}
\vspace{-2mm}
\end{figure}

\textbf{Nordland Railway}: The Nordland dataset \cite{IEEEexample:nordland} covers a 728km train journey from Trondheim to Bod\o\, in Nordland, Norway. This 10 hour train ride has been recorded four times, once per season: \textit{summer}, \textit{spring}, \textit{fall}, and \textit{winter}. Fig. \ref{nord-travs} shows a sample image for each traversal we used in our experiments, and Fig. \ref{gps}-left shows the related raw GPS data; which is more consistent compared to the Oxford RobotCar raw GPS data (see Fig. \ref{gps}-right).

\subsection{Motion Estimation}\label{sub:motion-estimation}

To provide our agent with motion data we separately use three different sensor modalities in our experiments: raw GPS data, visual odometry (VO), and optimized radar odometry (RO). For the Oxford RobotCar dataset, it already provides both GPS and VO sensor data. For RO, we used the optimized ground truth RO sensor data provided in the extended Oxford Radar RobotCar dataset \cite{IEEEexample:barnes2019oxford}---which has been demonstrated to be more accurate under challenging visual transitions---carefully chosen to visually match our selected traversals. For the Nordland dataset, however, we used the provided raw GPS data only as it is consistent across every traversal (Fig. \ref{gps}-left).

The goal of using these two datasets is to evaluate the effectiveness of our MVP-based approach in situations where vision-only navigation methods typically fail. For Nordland, when GPS data is fully available and consistent, we show that our approach can generalize better than visual-based navigation systems under extreme visual transitions. Similarly, for Oxford RobotCar, our method outperforms these visual-based navigation pipelines again under extreme visual changes, and also when GPS data reception fails.

\subsection{Visual Representations}\label{sub:visual-representations}

To enable sample-efficient RL training, as per previous work \cite{IEEEexample:chancan2020citylearn}, we encode all our full resolution RGB sensory images using the off-the-shelf VPR model NetVLAD; based on a VGG-16 network architecture \cite{IEEEexample:simonyan2014very} with PCA plus whitening on top their model. This deep-learning-based model is known to provide significantly better image representations compared to other VPR approaches \cite{IEEEexample:levelling}, and also enables to obtain compact feature dimensions (e.g., from 4096-\textit{d} all the way down to 64-\textit{d}). However, other deep-learning- or VPR-based models can equally used to encode our raw images. In this work, we used 64-\textit{d} image representations, $\mathbf{x}_t$, in all our MVP-based experiments. We then combine it with compact 2-\textit{d} motion representations, $\mathbf{m}_t$, to generate equally compact bimodal representations, $\mathbf{b}_t$, that feed our navigation policy network, see Fig. \ref{baselines} (a). We encoded $\mathbf{m}_t$ into compact feature vectors to preserve the compactness of $\mathbf{b}_t$, but it can be encoded using larger representations as in \cite{IEEEexample:mirowski2018learning}.

\subsection{MVP-based Policy Learning}\label{sub:mvp}

\begin{figure}[!ht]
\centering
\includegraphics[width=0.80\columnwidth]{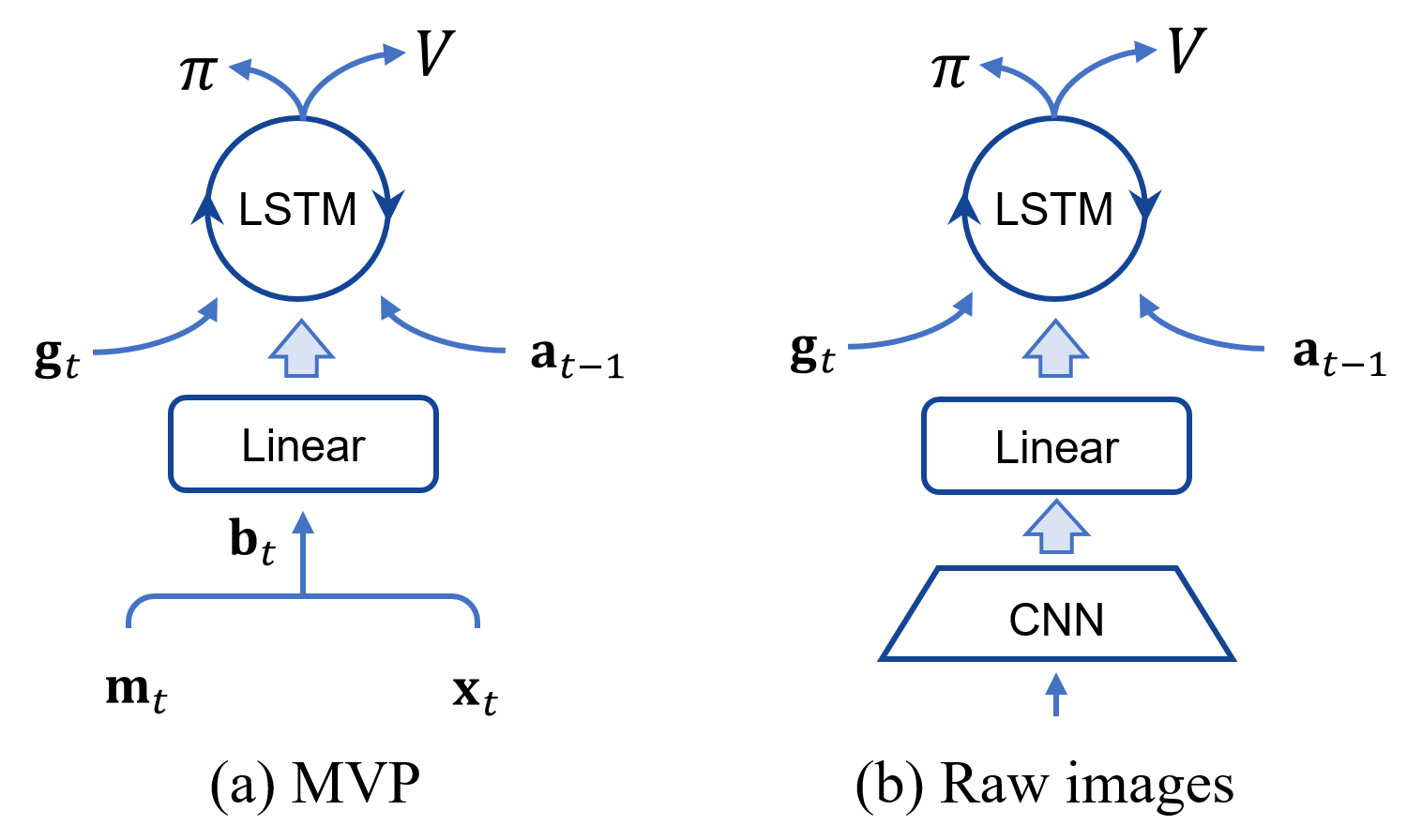}
\vspace{-2mm}
\caption{\textbf{Baselines agents}. We compared (a) our MVP-based approach to an (b) end-to-end vision-based policy network model based on raw images and also relies on GPS data for ground truth labeling as in \cite{IEEEexample:mirowski2018learning}.}
\label{baselines}
\end{figure}

\textbf{Goal-driven navigation}: Our method is trained on both motion ($\mathbf{m}_t$) and visual representations ($\mathbf{x}_t$) to successfully navigate through actions ($\mathbf{a}_t$) towards a required goal destination ($\mathbf{g}_t$), which is also encoded using 2-\textit{d} feature representations, over a single traversal in our CityLearn environment; see Fig. \ref{approach} and Fig. \ref{baselines} (a) for further details.

\textbf{Network Architecture}: We design our network model inspired by \cite{IEEEexample:mirowski2018learning}, see Fig. \ref{baselines} (a). A single \textit{linear} layer with $512$ units encodes our MVP bimodal representation ($\mathbf{b}_t$) to then combine it with the agent's previous actions ($\mathbf{a}_{t-1}$), using a single recurrent layer long short-term memory (LSTM) \cite{IEEEexample:lstm} with $256$ units. Updated agent's actions ($\mathbf{a}_{t}$) are also used to estimate both the required next actions and the value function $V$ from our policy network ($\pi_\theta$). To optimize $\pi_\theta$, we use the proximal policy optimization (PPO) algorithm \cite{IEEEexample:ppo}, which evaluates our objective function in Eq. \eqref{objective} for policy learning. We choose PPO as it can properly balance the small sample complexity of our compact input modalities and fine tuning requirements. 

\textbf{Reward design and curriculum learning}: 
We use multiple levels of curriculum learning \cite{IEEEexample:curriculumlearning} to gradually encourage our agent to explore the environment, and a sparse reward function that gives the agent a reward of $+1$ only when it finds the target.

\subsection{Vision-based Navigation Agent}\label{sub:vision-only}

We compare our MVP-based agent against a visual navigation agent with network architecture as proposed in \cite{IEEEexample:mirowski2018learning}, see Figs. \ref{baselines} (a) and (b).  This raw image based agent is adapted to also use a 2-\textit{d} feature vector for $\mathbf{g}_t$ to enable a fair comparison. In contrast to our method, this agent relies on GPS data for image labeling during both training and deployment and also does not incorporate motion learning. The network architecture of this agent includes a visual module of $2$ \textit{convolutional} layers, as per previous work \cite{IEEEexample:impala, IEEEexample:mirowski2016complex}, with RGB input images of $84\times84$ pixels. The first CNN layer uses a kernel of size $8\times8$, stride of $4\times4$, and 16 feature maps, and the second CNN layer uses a kernel of size $4\times4$, stride of $2\times2$, and 32 feature maps.

\subsection{Evaluation Metrics}\label{sub:metrics}

\begin{figure*}[!t]
\centering
\subfigure{\includegraphics[width=\columnwidth]{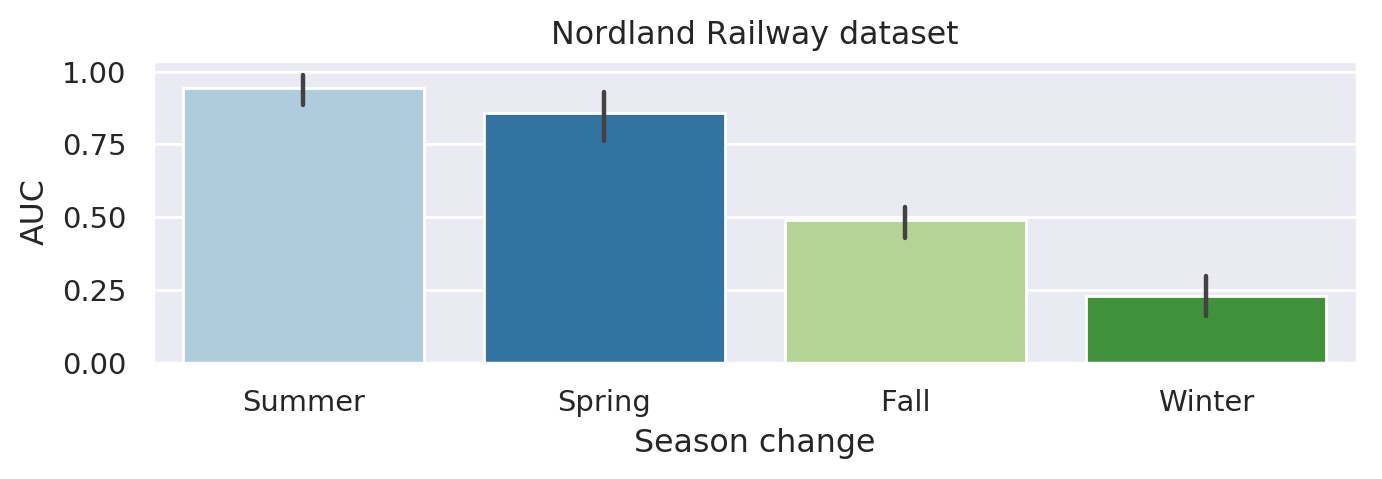}}
\subfigure{\includegraphics[width=\columnwidth]{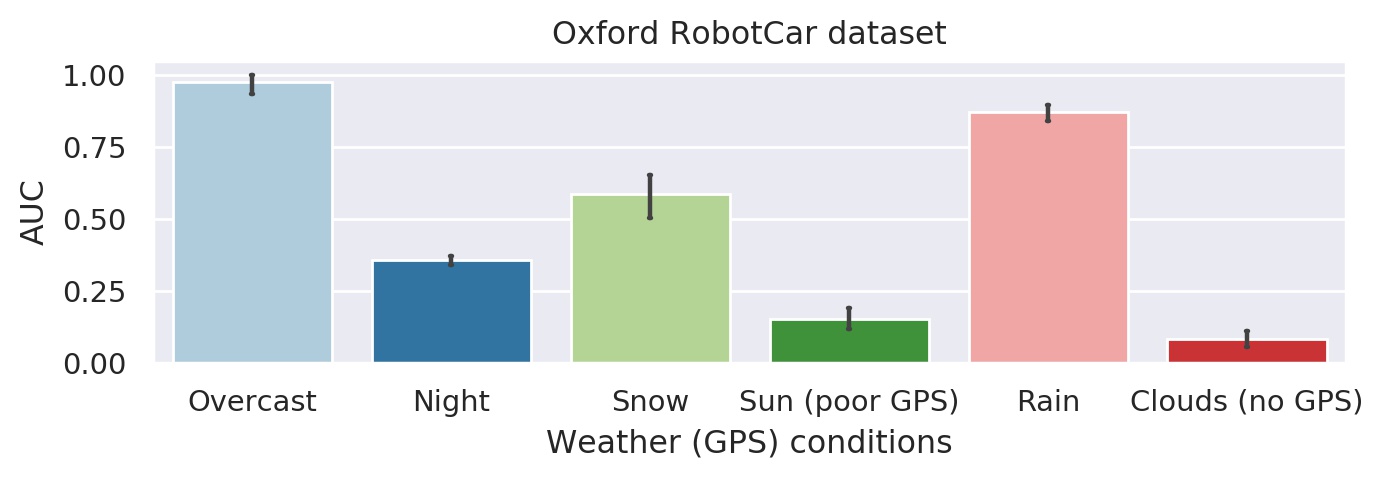}}
\vspace{-2mm}
\caption{\textbf{Visual place recognition experiments}. Conventional single-frame VPR results on Nordland Railway (left) and Oxford RobotCar (right) datasets. For Nordland, we show how our 64-\textit{d} visual representations perform under season changes. For Oxford Robotcar, we additionally show how VPR methods suffer when using poor GPS data as ground truth.}
\label{vpr-resuts}
\end{figure*}

\textbf{VPR experiments}: We report extensive VPR results, obtained using our compact image representations ($\mathbf{x}_{t}$), in order to provide an indicator of the visual component performance underlying our overall RL-based MVP system. A \textit{linear} classifier is trained on each reference traversal to then evaluate it on the remaining query traversals. Classification scores obtained for each image are then used to compute precision-recall curves, which are finally used to calculate our \textit{area under the curve} (AUC) results. AUC results on 10 experiments per traversal are presented in Fig. \ref{vpr-resuts}.

\begin{figure*}[!t]
\centering
\subfigure{\includegraphics[width=\columnwidth]{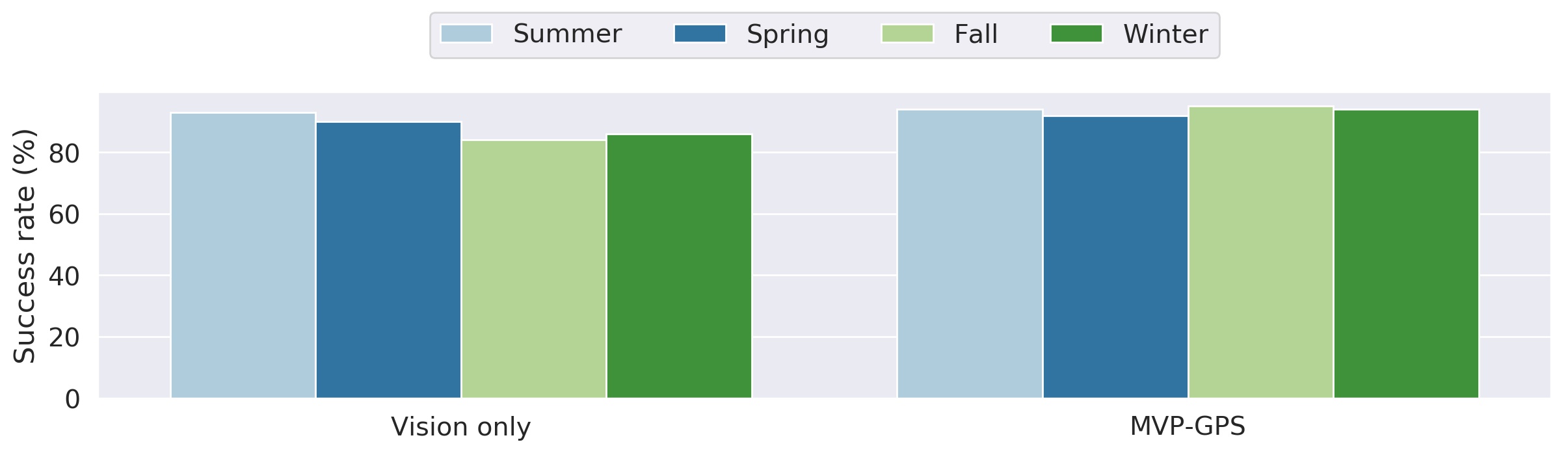}}
\subfigure{\includegraphics[width=\columnwidth]{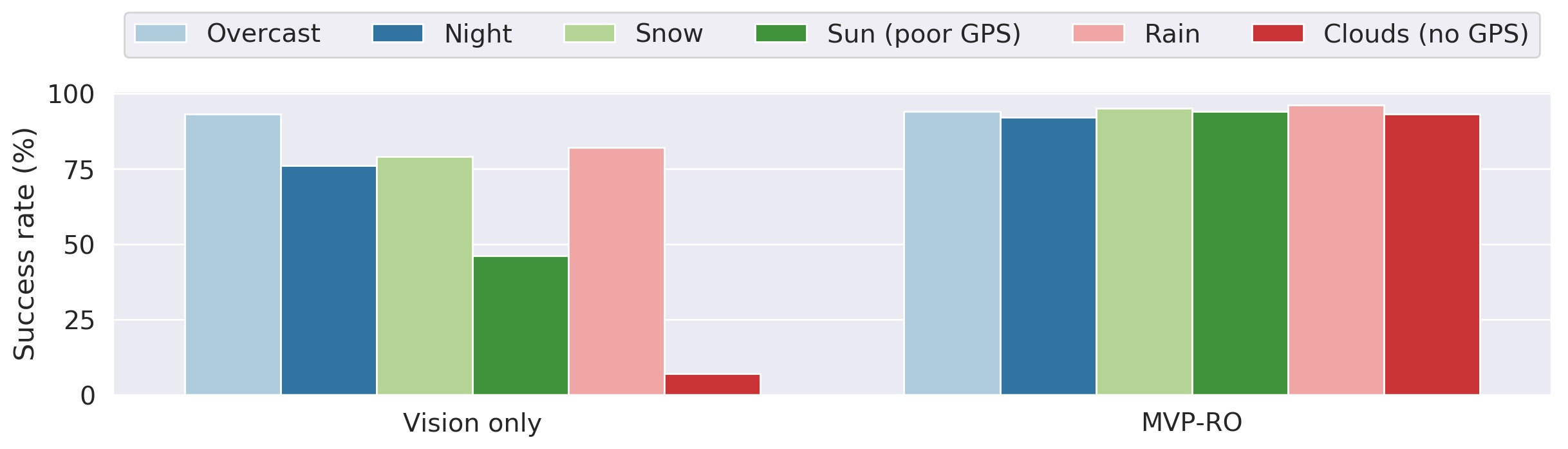}}
\vspace{-2mm}
\caption{\textbf{Reinforcement learning deployment}. Our MVP-based self-supervised method, that temporally aligns our compact visual representation with motion estimation data, achieves a 92\% navigation success rate across all-weather conditions on the Nordland (left) and the Oxford RobotCar (right) dataset, compared to vision-only navigation methods that do not to generalize well under extreme visual conditions (left) or rely on precise GPS data (right).}
\label{rl-results}
\end{figure*}

\textbf{RL-based navigation tasks}: We evaluate our trained agents on all corresponding dataset traversals, and provide statistics on the number of successful tasks in terms of the \textit{success rate} results over 10 deployment iterations, each iteration with 100 different targets. Average results on those evaluations are reported in Fig. \ref{rl-results}. We additionally constrain the maximum number of agent steps per episode to be less than the number of images within the traversal as in \cite{IEEEexample:chancan2020citylearn}. 

\section{Experiments: Results}\label{sec:experiments}

We present two main experimental results: conventional single-frame visual place recognition evaluations (Fig. \ref{vpr-resuts}), and reinforcement learning deployment of the navigation agents; both evaluated in our two datasets (Fig. \ref{rl-results}). We also provide details on the influence of incorporating motion data into our network architecture, and present selected illustrative results from our MVP method during deployment.

\subsection{Visual Place Recognition Results}\label{sub:vpr-results}

In Fig. \ref{vpr-resuts} we provide a full overview of our visual place recognition experiments, as described in Sections \ref{sub:visual-representations} and \ref{sub:metrics}, in terms of AUC performance. For the Nordland dataset (Fig. \ref{vpr-resuts}-left), we trained a \textit{linear} classified on the \textit{summer} traversal, and then evaluated it on \textit{spring}, \textit{fall} and \textit{winter} conditions. It is observed that extreme environmental changes such as those from \textit{fall} and \textit{winter} significantly reduce the results to around 0.50 and 0.25 AUC, respectively. It is worth noting, again, that each traversal of this dataset (and also of the Oxford RobotCar dataset) is a single sequence of images, meaning that we have used a single image from a particular place for training, and we do not use any data augmentation technique.

For the Oxford RobotCar dataset (Fig. \ref{vpr-resuts}-right), we trained a \textit{linear} classifier on the \textit{overcast} traversal, and evaluated it on the remaining traversals. AUC results in this dataset are relatively lower compared to those from the Nordland dataset; except for the \textit{rain} traversal that achieves around 0.80 AUC. This is mainly because those traversals include significant viewpoint changes, diverse environmental transitions, and also real GPS data. For the \textit{sun} and \textit{clouds} traversals, with around 0.12 and 0.05 AUC, with poor and no GPS data reception, respectively, the results obtained highlight the importance of having good GPS data for ground truth labeling; especially for VPR. In contrast, the \textit{night} and \textit{snow} traversals, with good GPS data, still present relatively good result around 0.30 and 0.55 AUC, respectively; regardless of their significantly different visual and lighting conditions, compared to the \textit{overcast} traversal.

\subsection{Navigation Policy Deployment}\label{sub:rl-deployment}

Navigation success rate results of our RL policies are reported on the Nordland (Fig. \ref{rl-results}-left) and Oxford RobotCar (Fig. \ref{rl-results}-right) datasets. We also compare our method to a vision-only approach, as described in Sections \ref{sub:mvp} and \ref{sub:vision-only}, respectively, using our CityLearn environment.

For the vision-based agent (referred as \textit{vision only} in Fig. \ref{rl-results}), which has been trained on raw images only, it is notable that the \textit{success rate} results are significantly better than those from our VPR experiments (Fig. \ref{vpr-resuts}), with over 84\% success rate for the Nordland dataset (Fig. \ref{rl-results}-left) and more than 46\% for the Oxford RobotCar dataset (Fig. \ref{rl-results}-right); except for the \textit{clouds} traversal which has no GPS data. Suggesting that the generalization capabilities of the whole RL-based systems is robust to environmental variations. However, this method still does not generalize well under different weather conditions with significant viewpoint changes and occlusions, as in the Oxford RobotCar dataset, especially with poor GPS data (see \textit{vision only} in \ref{rl-results}-right for the \textit{sun} and \textit{clouds} traversals). Suggesting that RL-based vision-only navigation methods that rely on precise GPS information are likely to fail when using poor motion estimation information.

In contrast, our MVP-based agents overcome the limitations of the VPR module, underlying the vision-only method, by temporally incorporating those visual representations with precise motion information into our navigation policy network using either GPS (when fully available) or odometry-based techniques, referred in Fig. \ref{rl-results} as \textit{MVP-GPS} and \textit{MVP-RO}, respectively. On the Nordland dataset (Fig. \ref{rl-results}-left), the \textit{vision only} agent achieves around 86\% success rate under challenging \textit{winter} conditions, compared to around 94\% for the MVP-GPS agent (see green bar in both cases). Similarly, on the Oxford RobotCar dataset (Fig. \ref{rl-results}-right), the MVP-RO agent achieves 93\% success rate under \textit{clouds} conditions, with no GPS data available, compared to 7\% for the \textit{vision only} agent (see red bar in both cases).

\subsection{Influence of Motion Estimation Precision}\label{sub:influence}

To analyze the influence of including motion data as an input to our policy network, we provide additional results on the Oxford RobotCar dataset shown in Fig. \ref{influence}. Vision-based navigation methods actually generalize relatively well under extreme changes with a 75\% success rate from day to night transitions, but with good GPS data reception. However, these methods can fail when GPS data is not precise, even under similar visual conditions, such as day to clouds (day) changes, with a 7\% success rate (see green bars for both cases in Fig. \ref{influence}-left). Conversely, our MVP-based approach leverages the use of relatively precise motion estimation data, including but not limited to those from radar or visual odometry, on top of those vision navigation methods to improve overall performance under both visual changes and when there is no GPS data available (see orange and blue bars in Fig. \ref{influence}-left). In Fig. \ref{influence}-right, we characterize the deployment performance of our MVP-based method using VO to estimate motion data. This graph shows how incorporating precise motion information can improve the overall navigation performance of our system, suggesting that current vision-only navigation methods can also benefit from using MVP-like approaches. As also demonstrated in related work \cite{IEEEexample:vo2004cvpr}, odometry-based techniques can be used directly for navigation tasks. This method, however, may require additional baseline metrics to estimate global scale factors during deployment on real robots; particularly when using relative motion data relative to the robot initial pose.

\subsection{Generalization Results}

\begin{figure}[!t]
\centering
\includegraphics[width=\columnwidth]{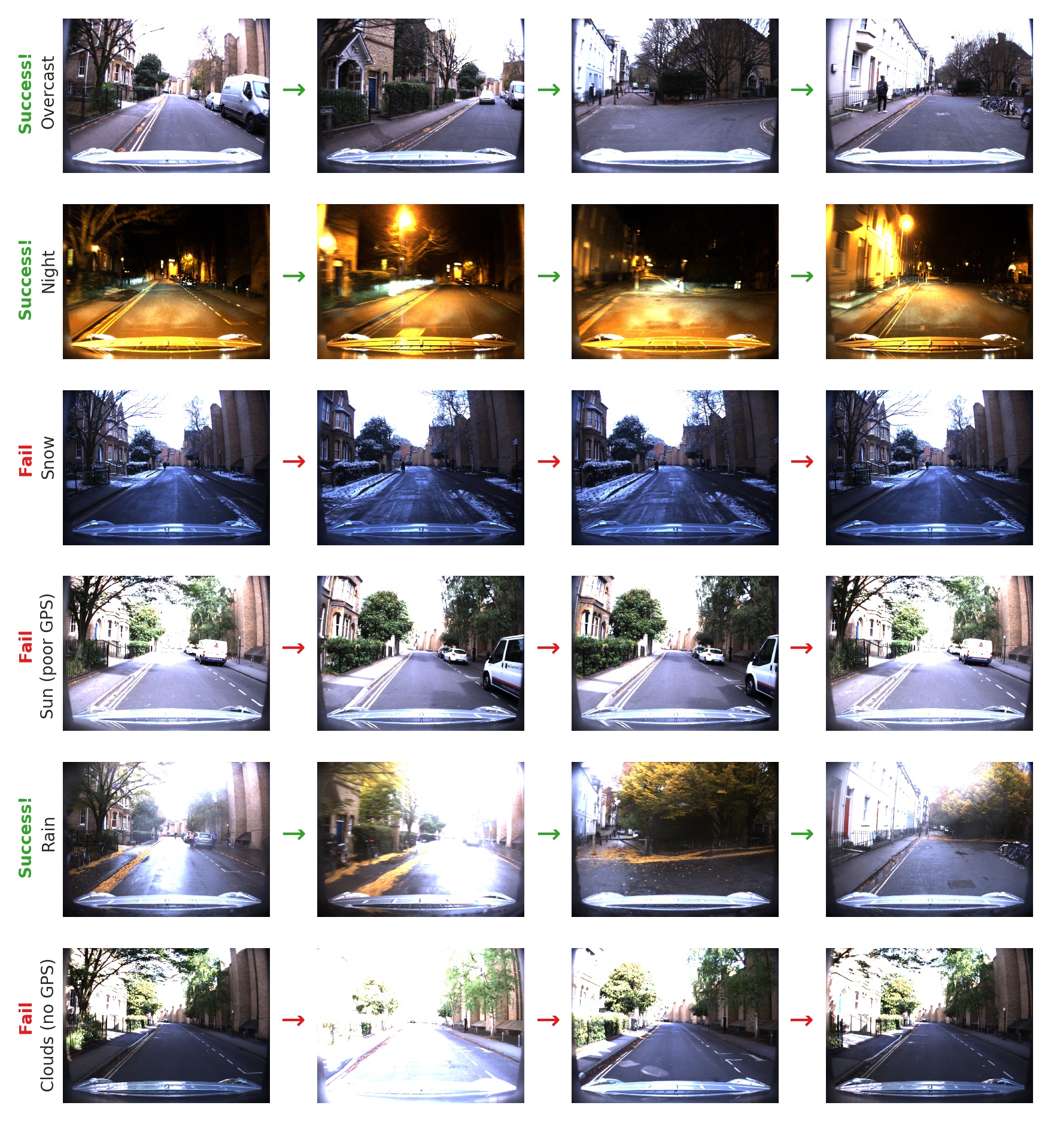}
\vspace{-6mm}
\caption{\textbf{Vision-based navigation results}. This approach can navigate to the target on \textit{overcast}, \textit{night} and \textit{rain}, but fails on the other traversals.}
\label{deploy-vis}
\end{figure}

\begin{figure}[!t]
\centering
\includegraphics[width=\columnwidth]{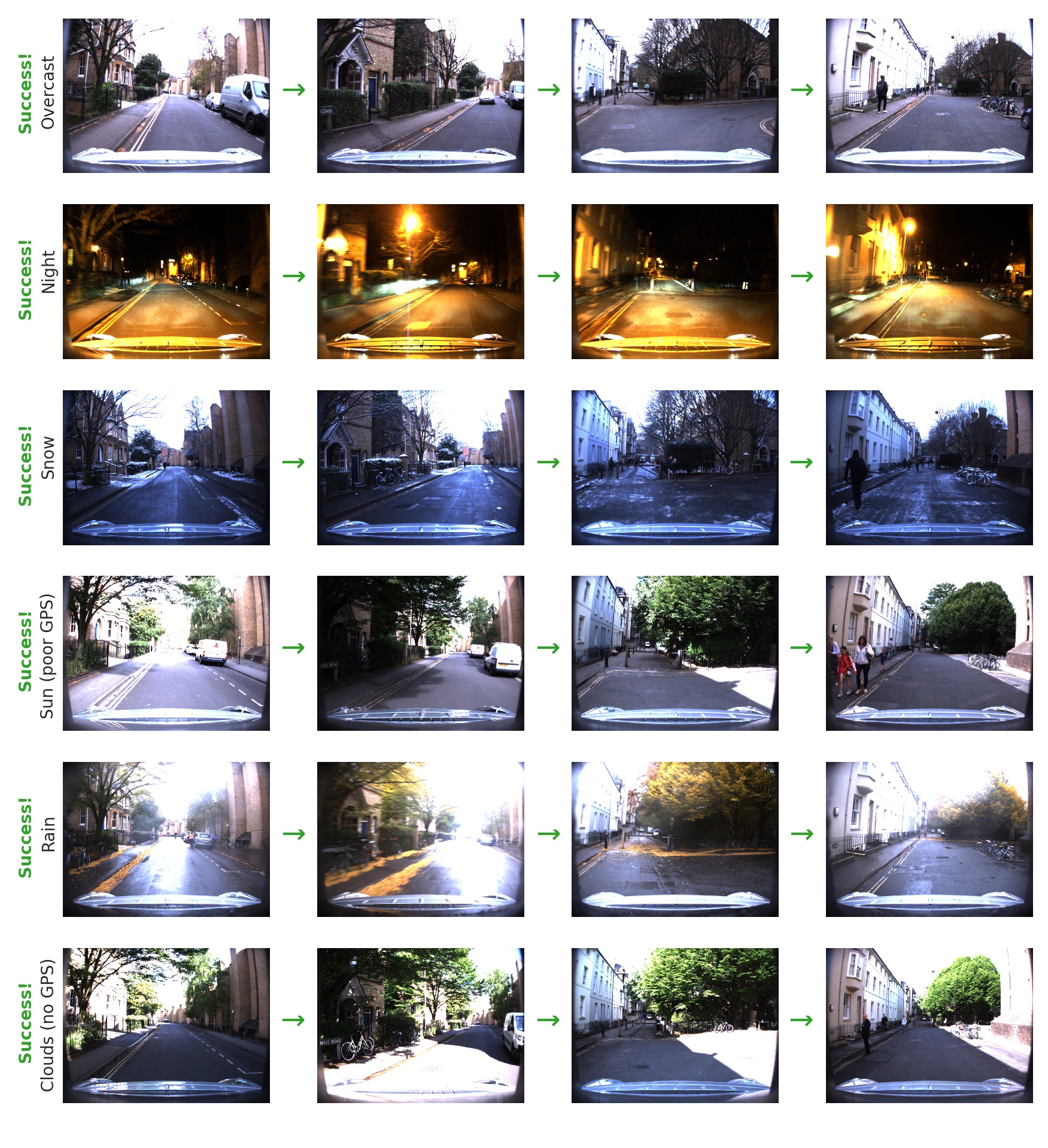}
\vspace{-6mm}
\caption{\textbf{MVP-based navigation results}. Our approach is capable of precisely navigating to the goal across all the traversals of the Oxford RobotCar dataset.}
\label{deploy-mvp}
\vspace{-2mm}
\end{figure}

We present illustrative navigation deployment results in Figs. \ref{deploy-vis} and \ref{deploy-mvp} for the vision-based agent and for our MVP-based approach, respectively. The agent is required to navigate from the same starting location to a distant target over all our selected traversals of the Oxford RobotCar dataset; see navigation states from left to right in Figs. \ref{deploy-vis} and \ref{deploy-mvp} including two intermediate states. Our approach is capable of precisely navigating to the target for every condition change (see Fig. \ref{deploy-mvp}), while the vision-based agent fails under extreme condition variations and also where GPS data is poor or not available (Fig. \ref{deploy-vis}).

\section{Conclusions}

We have proposed a method including a new network architecture that temporally integrate two fundamental sensor modalities such as motion and visual perception (MVP) information for large-scale target-driven navigation tasks using real data via reinforcement learning (RL). Our MVP-based approach was demonstrated to be robust to both extreme visual changing conditions and also poor absolute positioning information such as those from GPS, where typical visual (only) navigation pipelines fail. This suggests that the incorporation of motion information, including but not limited to GPS (when fully available) or visual/radar odometry, can be used to improve the overall performance and robustness of conventional visual-based navigation systems that rely on raw images only for learning complex navigation tasks. Future work combining different motion estimation modalities such as linear/angular velocities with visual representations is likely to be considered. However, this could potentially increase the network complexity and training requirements \cite{IEEEexample:banino2018vector,IEEEexample:cueva2018emergence}, especially when using real data. Quantifying the relationship between required RL performance, visual place recognition generalization capabilities, and motion estimation quality can also provide new insights for selecting between different motion estimation sensor modalities for a specific robotic navigation system.

\bibliographystyle{IEEEtran}
\bibliography{IEEEexample}

\begin{thebibliography}{10}
\providecommand{\url}[1]{#1}
\csname url@rmstyle\endcsname
\providecommand{\newblock}{\relax}
\providecommand{\bibinfo}[2]{#2}
\providecommand\BIBentrySTDinterwordspacing{\spaceskip=0pt\relax}
\providecommand\BIBentryALTinterwordstretchfactor{4}
\providecommand\BIBentryALTinterwordspacing{\spaceskip=\fontdimen2\font plus
\BIBentryALTinterwordstretchfactor\fontdimen3\font minus
  \fontdimen4\font\relax}
\providecommand\BIBforeignlanguage[2]{{%
\expandafter\ifx\csname l@#1\endcsname\relax
\typeout{** WARNING: IEEEtran.bst: No hyphenation pattern has been}%
\typeout{** loaded for the language `#1'. Using the pattern for}%
\typeout{** the default language instead.}%
\else
\language=\csname l@#1\endcsname
\fi
#2}}

\bibitem{IEEEexample:mirowski2018learning}
{P. Mirowski \textit{et al.}}, ``Learning to navigate in cities without a
  map,'' in \emph{Advances in Neural Information Processing Systems}, 2018, pp.
  2419--2430.

\bibitem{IEEEexample:ma2019towards}
{H. Ma \textit{et al.}}, ``Towards navigation without precise localization:
  Weakly supervised learning of goal-directed navigation cost map,''
  \emph{arXiv preprint arXiv:1906.02468}, 2019.

\bibitem{IEEEexample:streetinst}
{V. Cirik \textit{et al.}}, ``Following formulaic map instructions in a street
  simulation environment,'' in \emph{2018 NeurIPS Workshop on Visually Grounded
  Interaction and Language}, vol.~1, 2018.

\bibitem{IEEEexample:hermann2019learning}
K.~M. Hermann, M.~Malinowski, P.~Mirowski, A.~Banki-Horvath, K.~Anderson, and
  R.~Hadsell, ``Learning to follow directions in street view,'' \emph{arXiv
  preprint arXiv:1903.00401}, 2019.

\bibitem{IEEEexample:Chen_2019_CVPR}
H.~Chen, A.~Suhr, D.~Misra, N.~Snavely, and Y.~Artzi, ``Touchdown: Natural
  language navigation and spatial reasoning in visual street environments,'' in
  \emph{The IEEE Conference on Computer Vision and Pattern Recognition (CVPR)},
  June 2019.

\bibitem{IEEEexample:talkwalk}
H.~de~Vries, K.~Shuster, D.~Batra, D.~Parikh, J.~Weston, and D.~Kiela, ``Talk
  the walk: Navigating new york city through grounded dialogue,'' \emph{arXiv
  preprint arXiv:1807.03367}, 2018.

\bibitem{IEEEexample:chancan2020citylearn}
M.~Chanc{\'a}n and M.~Milford, ``{From Visual Place Recognition to Navigation:
  Learning Sample-Efficient Control Policies across Diverse Real World
  Environments},'' \emph{arXiv preprint arXiv:1910.04335}, 2019, accepted to
  ICRA 2020.

\bibitem{IEEEexample:mohamed2019surveyodometry}
S.~A.~S. {Mohamed}, M.~{Haghbayan}, T.~{Westerlund}, J.~{Heikkonen},
  H.~{Tenhunen}, and J.~{Plosila}, ``A survey on odometry for autonomous
  navigation systems,'' \emph{IEEE Access}, vol.~7, pp. 97\,466--97\,486, 2019.

\bibitem{IEEEexample:barnes2019oxford}
D.~Barnes, M.~Gadd, P.~Murcutt, P.~Newman, and I.~Posner, ``{The Oxford Radar
  RobotCar Dataset: A Radar Extension to the Oxford RobotCar Dataset},''
  \emph{arXiv preprint arXiv:1909.01300}, 2019.

\bibitem{IEEEexample:netvlad}
R.~Arandjelovi{\'c}, P.~Gron{\'a}t, A.~Torii, T.~Pajdla, and J.~Sivic,
  ``{NetVLAD: CNN Architecture for Weakly Supervised Place Recognition},''
  \emph{2016 IEEE Conference on Computer Vision and Pattern Recognition
  (CVPR)}, pp. 5297--5307, 2016.

\bibitem{IEEEexample:maddern2017oxford}
W.~Maddern, G.~Pascoe, C.~Linegar, and P.~Newman, ``1 year, 1000 km: The oxford
  robotcar dataset,'' \emph{The International Journal of Robotics Research},
  vol.~36, no.~1, pp. 3--15, 2017.

\bibitem{IEEEexample:nordland}
N.~S{\"u}nderhauf, P.~Neubert, and P.~Protzel, ``Are we there yet? challenging
  seqslam on a 3000 km journey across all four seasons,'' in \emph{Proc.
  Workshop Long-Term Autonomy 2013 IEEE Int. Conf. Robot. Autom. (ICRA)}, 2013.

\bibitem{IEEEexample:cadena2016slam}
C.~{Cadena}, L.~{Carlone}, H.~{Carrillo}, Y.~{Latif}, D.~{Scaramuzza},
  J.~{Neira}, I.~{Reid}, and J.~J. {Leonard}, ``Past, present, and future of
  simultaneous localization and mapping: Toward the robust-perception age,''
  \emph{IEEE Transactions on Robotics}, vol.~32, no.~6, pp. 1309--1332, Dec
  2016.

\bibitem{IEEEexample:vo2004cvpr}
D.~Nist\'er, O.~{Naroditsky}, and J.~{Bergen}, ``Visual odometry,'' in
  \emph{Proceedings of the 2004 IEEE Computer Society Conference on Computer
  Vision and Pattern Recognition (CVPR)}, vol.~1, June 2004.

\bibitem{IEEEexample:dissanayake2001radar}
M.~W. M.~G. {Dissanayake}, P.~{Newman}, S.~{Clark}, H.~F. {Durrant-Whyte}, and
  M.~{Csorba}, ``{A solution to the simultaneous localization and map building
  (SLAM) problem},'' \emph{IEEE Transactions on Robotics and Automation},
  vol.~17, no.~3, pp. 229--241, June 2001.

\bibitem{IEEEexample:davison2007monoslam}
A.~J. {Davison}, I.~D. {Reid}, N.~D. {Molton}, and O.~{Stasse}, ``{MonoSLAM:
  Real-Time Single Camera SLAM},'' \emph{IEEE Transactions on Pattern Analysis
  and Machine Intelligence}, vol.~29, no.~6, pp. 1052--1067, June 2007.

\bibitem{IEEEexample:orbslam}
R.~{Mur-Artal}, J.~M.~M. {Montiel}, and J.~D. {Tard\'os}, ``{ORB-SLAM: A
  Versatile and Accurate Monocular SLAM System},'' \emph{IEEE Transactions on
  Robotics}, vol.~31, no.~5, pp. 1147--1163, Oct 2015.

\bibitem{IEEEexample:ratslam2004}
M.~J. {Milford}, G.~F. {Wyeth}, and D.~{Prasser}, ``Ratslam: a hippocampal
  model for simultaneous localization and mapping,'' in \emph{IEEE
  International Conference on Robotics and Automation (ICRA)}, vol.~1, April
  2004, pp. 403--408.

\bibitem{IEEEexample:ratslam2008}
M.~J. {Milford} and G.~F. {Wyeth}, ``Mapping a suburb with a single camera
  using a biologically inspired slam system,'' \emph{IEEE Transactions on
  Robotics}, vol.~24, no.~5, pp. 1038--1053, Oct 2008.

\bibitem{IEEEexample:wifi2010}
J.~{Biswas} and M.~{Veloso}, ``Wifi localization and navigation for autonomous
  indoor mobile robots,'' in \emph{2010 IEEE International Conference on
  Robotics and Automation}, May 2010, pp. 4379--4384.

\bibitem{IEEEexample:renato2016}
R.~{Miyagusuku}, A.~{Yamashita}, and H.~{Asama}, ``Improving gaussian processes
  based mapping of wireless signals using path loss models,'' in \emph{2016
  IEEE/RSJ International Conference on Intelligent Robots and Systems (IROS)},
  Oct 2016, pp. 4610--4615.

\bibitem{IEEEexample:renato2017}
R.~{Miyagusuku}, Y.~{Seow}, A.~{Yamashita}, and H.~{Asama}, ``Fast and robust
  localization using laser rangefinder and wifi data,'' in \emph{2017 IEEE
  International Conference on Multisensor Fusion and Integration for
  Intelligent Systems (MFI)}, Nov 2017, pp. 111--117.

\bibitem{IEEEexample:renato2019ral}
R.~{Miyagusuku}, A.~{Yamashita}, and H.~{Asama}, ``{Data Information Fusion
  From Multiple Access Points for WiFi-Based Self-localization},'' \emph{IEEE
  Robotics and Automation Letters}, vol.~4, no.~2, pp. 269--276, April 2019.

\bibitem{IEEEexample:orbslam2}
R.~{Mur-Artal} and J.~D. {Tard\'os}, ``{ORB-SLAM2: An Open-Source SLAM System
  for Monocular, Stereo, and RGB-D Cameras},'' \emph{IEEE Transactions on
  Robotics}, vol.~33, no.~5, pp. 1255--1262, Oct 2017.

\bibitem{IEEEexample:muratal2017vimslam}
R.~{Mur-Artal} and J.~D. Tard\'os, ``{Visual-Inertial Monocular SLAM With Map
  Reuse},'' \emph{IEEE Robotics and Automation Letters}, vol.~2, no.~2, pp.
  796--803, April 2017.

\bibitem{IEEEexample:vo}
D.~Nist\'er, O.~Naroditsky, and J.~Bergen, ``Visual odometry for ground vehicle
  applications,'' \emph{Journal of Field Robotics}, vol.~23, no.~1, pp. 3--20,
  2006.

\bibitem{IEEEexample:doh2006relative}
N.~L. Doh, H.~Choset, and W.~K. Chung, ``Relative localization using path
  odometry information,'' \emph{Autonomous Robots}, vol.~21, no.~2, pp.
  143--154, 2006.

\bibitem{IEEEexample:pascoe2017nid}
G.~Pascoe, W.~Maddern, M.~Tanner, P.~Pini{\'e}s, and P.~Newman, ``Nid-slam:
  Robust monocular slam using normalised information distance,'' in
  \emph{Proceedings of the IEEE Conference on Computer Vision and Pattern
  Recognition}, 2017, pp. 1435--1444.

\bibitem{IEEEexample:kim2018lowdriftvo}
P.~{Kim}, B.~{Coltin}, and H.~J. {Kim}, ``Low-drift visual odometry in
  structured environments by decoupling rotational and translational motion,''
  in \emph{2018 IEEE International Conference on Robotics and Automation
  (ICRA)}, May 2018, pp. 7247--7253.

\bibitem{IEEEexample:zhan2019visual}
H.~Zhan, C.~S. Weerasekera, J.~Bian, and I.~Reid, ``Visual odometry revisited:
  What should be learnt?'' \emph{arXiv preprint arXiv:1909.09803}, 2019.

\bibitem{IEEEexample:svo}
C.~{Forster}, Z.~{Zhang}, M.~{Gassner}, M.~{Werlberger}, and D.~{Scaramuzza},
  ``{SVO: Semidirect Visual Odometry for Monocular and Multicamera Systems},''
  \emph{IEEE Transactions on Robotics}, vol.~33, no.~2, pp. 249--265, April
  2017.

\bibitem{IEEEexample:hong2017vio}
E.~{Hong} and J.~{Lim}, ``Visual inertial odometry using coupled nonlinear
  optimization,'' in \emph{2017 IEEE/RSJ International Conference on
  Intelligent Robots and Systems (IROS)}, Sep. 2017, pp. 6879--6885.

\bibitem{IEEEexample:babu2018vio}
B.~P.~W. {Babu}, D.~{Cyganski}, J.~{Duckworth}, and S.~{Kim}, ``Detection and
  resolution of motion conflict in visual inertial odometry,'' in \emph{2018
  IEEE International Conference on Robotics and Automation (ICRA)}, May 2018,
  pp. 996--1002.

\bibitem{IEEEexample:zhang2015visionlidar}
J.~{Zhang} and S.~{Singh}, ``Visual-lidar odometry and mapping: low-drift,
  robust, and fast,'' in \emph{2015 IEEE International Conference on Robotics
  and Automation (ICRA)}, May 2015, pp. 2174--2181.

\bibitem{IEEEexample:borges2018posemap}
P.~{Egger}, P.~V.~K. {Borges}, G.~{Catt}, A.~{Pfrunder}, R.~{Siegwart}, and
  R.~{Dubé}, ``Posemap: Lifelong, multi-environment 3d lidar localization,''
  in \emph{2018 IEEE/RSJ International Conference on Intelligent Robots and
  Systems (IROS)}, Oct 2018, pp. 3430--3437.

\bibitem{IEEEexample:wang2019vilaser}
Z.~{Wang}, J.~{Zhang}, S.~{Chen}, C.~{Yuan}, J.~{Zhang}, and J.~{Zhang},
  ``Robust high accuracy visual-inertial-laser slam system,'' in \emph{2019
  IEEE/RSJ International Conference on Intelligent Robots and Systems (IROS)},
  Nov 2019, pp. 6636--6641.

\bibitem{IEEEexample:cen2018radar}
S.~H. {Cen} and P.~{Newman}, ``Precise ego-motion estimation with
  millimeter-wave radar under diverse and challenging conditions,'' in
  \emph{2018 IEEE International Conference on Robotics and Automation (ICRA)},
  May 2018, pp. 6045--6052.

\bibitem{IEEEexample:wang2017learnVO}
S.~{Wang}, R.~{Clark}, H.~{Wen}, and N.~{Trigoni}, ``Deepvo: Towards end-to-end
  visual odometry with deep recurrent convolutional neural networks,'' in
  \emph{2017 IEEE International Conference on Robotics and Automation (ICRA)},
  May 2017, pp. 2043--2050.

\bibitem{IEEEexample:pillai2017learnvo}
S.~{Pillai} and J.~J. {Leonard}, ``Towards visual ego-motion learning in
  robots,'' in \emph{2017 IEEE/RSJ International Conference on Intelligent
  Robots and Systems (IROS)}, Sep. 2017, pp. 5533--5540.

\bibitem{IEEEexample:zhou2017learnvo}
T.~Zhou, M.~Brown, N.~Snavely, and D.~G. Lowe, ``Unsupervised learning of depth
  and ego-motion from video,'' in \emph{The IEEE Conference on Computer Vision
  and Pattern Recognition (CVPR)}, July 2017.

\bibitem{IEEEexample:zhan2018unsupervised}
H.~Zhan, R.~Garg, C.~Saroj~Weerasekera, K.~Li, H.~Agarwal, and I.~Reid,
  ``Unsupervised learning of monocular depth estimation and visual odometry
  with deep feature reconstruction,'' in \emph{Proceedings of the IEEE
  Conference on Computer Vision and Pattern Recognition}, 2018, pp. 340--349.

\bibitem{IEEEexample:Casser_2019_CVPR_Workshops}
V.~Casser, S.~Pirk, R.~Mahjourian, and A.~Angelova, ``Unsupervised monocular
  depth and ego-motion learning with structure and semantics,'' in \emph{The
  IEEE Conference on Computer Vision and Pattern Recognition (CVPR) Workshops},
  June 2019.

\bibitem{IEEEexample:wang2019learnvo}
{X. {Wang} \textit{et al.}}, ``Improving learning-based ego-motion estimation
  with homomorphism-based losses and drift correction,'' in \emph{IEEE/RSJ
  International Conference on Intelligent Robots and Systems (IROS)}, 2019, pp.
  970--976.

\bibitem{IEEEexample:loo2019cnn-vo}
S.~Y. {Loo}, A.~J. {Amiri}, S.~{Mashohor}, S.~H. {Tang}, and H.~{Zhang},
  ``Cnn-svo: Improving the mapping in semi-direct visual odometry using
  single-image depth prediction,'' in \emph{2019 International Conference on
  Robotics and Automation (ICRA)}, May 2019, pp. 5218--5223.

\bibitem{IEEEexample:shen2019learnvo}
T.~{Shen}, Z.~{Luo}, L.~{Zhou}, H.~{Deng}, R.~{Zhang}, T.~{Fang}, and
  L.~{Quan}, ``Beyond photometric loss for self-supervised ego-motion
  estimation,'' in \emph{2019 International Conference on Robotics and
  Automation (ICRA)}, May 2019, pp. 6359--6365.

\bibitem{IEEEexample:Shen_2019_ICCV}
T.~Shen, L.~Zhou, Z.~Luo, Y.~Yao, S.~Li, J.~Zhang, T.~Fang, and L.~Quan,
  ``Self-supervised learning of depth and motion under photometric
  inconsistency,'' in \emph{The IEEE International Conference on Computer
  Vision (ICCV) Workshops}, Oct 2019.

\bibitem{IEEEexample:suaftescu2020kidnapped}
{\c{S}}.~S{\u{a}}ftescu, M.~Gadd, D.~De~Martini, D.~Barnes, and P.~Newman,
  ``Kidnapped radar: Topological radar localisation using
  rotationally-invariant metric learning,'' \emph{arXiv preprint
  arXiv:2001.09438}, 2020.

\bibitem{IEEEexample:Torii_2015_CVPR}
A.~Torii, R.~Arandjelovic, J.~Sivic, M.~Okutomi, and T.~Pajdla, ``24/7 place
  recognition by view synthesis,'' in \emph{The IEEE Conference on Computer
  Vision and Pattern Recognition (CVPR)}, June 2015.

\bibitem{IEEEexample:weyand2016planet}
T.~Weyand, I.~Kostrikov, and J.~Philbin, ``Planet-photo geolocation with
  convolutional neural networks,'' in \emph{European Conference on Computer
  Vision}.\hskip 1em plus 0.5em minus 0.4em\relax Springer, 2016, pp. 37--55.

\bibitem{IEEEexample:2017geo}
H.~J. {Kim}, E.~{Dunn}, and J.~{Frahm}, ``Learned contextual feature
  reweighting for image geo-localization,'' in \emph{2017 IEEE Conference on
  Computer Vision and Pattern Recognition (CVPR)}, July 2017, pp. 3251--3260.

\bibitem{IEEEexample:gordo2017end}
A.~Gordo, J.~Almazan, J.~Revaud, and D.~Larlus, ``End-to-end learning of deep
  visual representations for image retrieval,'' \emph{International Journal of
  Computer Vision}, vol. 124, no.~2, pp. 237--254, 2017.

\bibitem{IEEEexample:fine2019imret}
F.~Radenovi\'c, G.~Tolias, and O.~Chum, ``Fine-tuning cnn image retrieval with
  no human annotation,'' \emph{IEEE Transactions on Pattern Analysis and
  Machine Intelligence}, vol.~41, no.~7, pp. 1655--1668, July 2019.

\bibitem{IEEEexample:seqslam}
M.~J. {Milford} and G.~F. {Wyeth}, ``{SeqSLAM: Visual route-based navigation
  for sunny summer days and stormy winter nights},'' in \emph{2012 IEEE
  International Conference on Robotics and Automation (ICRA)}, May 2012, pp.
  1643--1649.

\bibitem{IEEEexample:pepperell2014all}
E.~Pepperell, P.~I. Corke, and M.~J. Milford, ``{All-environment visual place
  recognition with SMART},'' in \emph{2014 IEEE international conference on
  robotics and automation (ICRA)}, 2014, pp. 1612--1618.

\bibitem{IEEEexample:cnnlanmark}
{N. S{\"u}nderhauf \textit{et al.}}, ``Place recognition with convnet
  landmarks: Viewpoint-robust, condition-robust, training-free,''
  \emph{Proceedings of Robotics: Science and Systems XII}, 2015.

\bibitem{IEEEexample:sunderhauf2015performance}
{N. S{\"u}nderhauf\textit{et al.}}, ``On the performance of convnet features
  for place recognition,'' in \emph{2015 IEEE/RSJ International Conference on
  Intelligent Robots and Systems (IROS)}, 2015, pp. 4297--4304.

\bibitem{IEEEexample:mpf}
S.~Hausler, A.~Jacobson, and M.~Milford, ``Multi-process fusion: Visual place
  recognition using multiple image processing methods,'' \emph{IEEE Robotics
  and Automation Letters}, vol.~4, no.~2, pp. 1924--1931, 2019.

\bibitem{IEEEexample:geiger2012kitti}
A.~{Geiger}, P.~{Lenz}, and R.~{Urtasun}, ``{Are we ready for autonomous
  driving? The KITTI vision benchmark suite},'' in \emph{2012 IEEE Conference
  on Computer Vision and Pattern Recognition}, June 2012, pp. 3354--3361.

\bibitem{IEEEexample:guo2018safe}
J.~Guo, U.~Kurup, and M.~Shah, ``Is it safe to drive? an overview of factors,
  challenges, and datasets for driveability assessment in autonomous driving,''
  \emph{arXiv preprint arXiv:1811.11277}, 2018.

\bibitem{IEEEexample:naseer}
T.~{Naseer}, W.~{Burgard}, and C.~{Stachniss}, ``Robust visual localization
  across seasons,'' \emph{IEEE Transactions on Robotics}, vol.~34, no.~2, pp.
  289--302, April 2018.

\bibitem{IEEEexample:fabrat}
A.~J. {Glover}, W.~P. {Maddern}, M.~J. {Milford}, and G.~F. {Wyeth}, ``{FAB-MAP
  + RatSLAM: Appearance-based SLAM for multiple times of day},'' in \emph{2010
  IEEE International Conference on Robotics and Automation}, May 2010, pp.
  3507--3512.

\bibitem{IEEEexample:kitti}
A.~Geiger, P.~Lenz, C.~Stiller, and R.~Urtasun, ``{Vision meets robotics: The
  KITTI dataset},'' \emph{The International Journal of Robotics Research},
  vol.~32, no.~11, pp. 1231--1237, 2013.

\bibitem{IEEEexample:caesar2019nuscenes}
{H. Caesar \textit{et al.}}, ``{nuScenes: A multimodal dataset for autonomous
  driving},'' \emph{arXiv preprint arXiv:1903.11027}, 2019.

\bibitem{IEEEexample:vprsurvey}
{S. {Lowry} \textit{et al.}}, ``Visual place recognition: A survey,''
  \emph{IEEE Transactions on Robotics}, vol.~32, no.~1, pp. 1--19, Feb 2016.

\bibitem{IEEEexample:chen2014convolutional}
{Z. Chen \textit{et al}}, ``Convolutional neural network-based place
  recognition,'' \emph{arXiv preprint arXiv:1411.1509}, 2014.

\bibitem{IEEEexample:zetao2017}
{Z. Chen \textit{et al.}}, ``Deep learning features at scale for visual place
  recognition,'' \emph{2017 IEEE International Conference on Robotics and
  Automation (ICRA)}, pp. 3223--3230, 2017.

\bibitem{IEEEexample:lost}
S.~Garg, N.~S{\"u}enderhauf, and M.~Milford, ``Lost? appearance-invariant place
  recognition for opposite viewpoints using visual semantics,''
  \emph{Proceedings of Robotics: Science and Systems XIV}, 2018.

\bibitem{IEEEexample:chancan2020hybrid}
M.~{Chanc\'an}, L.~{Hernandez-Nunez}, A.~{Narendra}, A.~B. {Barron}, and
  M.~{Milford}, ``A hybrid compact neural architecture for visual place
  recognition,'' \emph{IEEE Robotics and Automation Letters}, vol.~5, no.~2,
  pp. 993--1000, April 2020.

\bibitem{IEEEexample:kahn2018composable}
G.~Kahn, A.~Villaflor, P.~Abbeel, and S.~Levine, ``Composable
  action-conditioned predictors: Flexible off-policy learning for robot
  navigation,'' in \emph{Proceedings of the 2nd Annual Conference on Robot
  Learning}, 2018, pp. 806--816.

\bibitem{IEEEexample:mirowski2016complex}
{Piotr Mirowski \textit{et al.}}, ``Learning to navigate in complex
  environments,'' \emph{arXiv preprint arXiv:1611.03673}, 2016.

\bibitem{IEEEexample:chaplot2018active}
D.~S. Chaplot, E.~Parisotto, and R.~Salakhutdinov, ``Active neural
  localization,'' \emph{arXiv preprint arXiv:1801.08214}, 2018.

\bibitem{IEEEexample:zhang2017neural}
J.~Zhang, L.~Tai, J.~Boedecker, W.~Burgard, and M.~Liu, ``Neural slam: Learning
  to explore with external memory,'' \emph{arXiv preprint arXiv:1706.09520},
  2017.

\bibitem{IEEEexample:khan2018drlnav}
G.~{Kahn}, A.~{Villaflor}, B.~{Ding}, P.~{Abbeel}, and S.~{Levine},
  ``Self-supervised deep reinforcement learning with generalized computation
  graphs for robot navigation,'' in \emph{2018 IEEE International Conference on
  Robotics and Automation (ICRA)}, May 2018, pp. 5129--5136.

\bibitem{IEEEexample:oh2019learnaction}
C.~{Oh} and A.~{Cavallaro}, ``Learning action representations for
  self-supervised visual exploration,'' in \emph{2019 International Conference
  on Robotics and Automation (ICRA)}, May 2019, pp. 5873--5879.

\bibitem{IEEEexample:kahn2020badgr}
G.~Kahn, P.~Abbeel, and S.~Levine, ``{BADGR: An Autonomous Self-Supervised
  Learning-Based Navigation System},'' \emph{arXiv preprint arXiv:2002.05700},
  2020.

\bibitem{IEEEexample:gupta2017cognitive}
S.~Gupta, J.~Davidson, S.~Levine, R.~Sukthankar, and J.~Malik, ``Cognitive
  mapping and planning for visual navigation,'' in \emph{Proceedings of the
  IEEE Conference on Computer Vision and Pattern Recognition}, 2017, pp.
  2616--2625.

\bibitem{IEEEexample:gupta2017unifying}
S.~Gupta, D.~Fouhey, S.~Levine, and J.~Malik, ``Unifying map and landmark based
  representations for visual navigation,'' \emph{arXiv preprint
  arXiv:1712.08125}, 2017.

\bibitem{IEEEexample:chen2019learning}
T.~Chen, S.~Gupta, and A.~Gupta, ``Learning exploration policies for
  navigation,'' \emph{arXiv preprint arXiv:1903.01959}, 2019.

\bibitem{IEEEexample:chaplot2020Learning}
D.~S. Chaplot, D.~Gandhi, S.~Gupta, A.~Gupta, and R.~Salakhutdinov, ``Learning
  to explore using active neural slam,'' in \emph{International Conference on
  Learning Representations}, 2020.

\bibitem{IEEEexample:banino2018vector}
{A. Banino \textit{et al.}}, ``Vector-based navigation using grid-like
  representations in artificial agents,'' \emph{Nature}, vol. 557, no. 7705,
  pp. 429--433, 2018.

\bibitem{IEEEexample:guo2014atari}
X.~Guo, S.~Singh, H.~Lee, R.~L. Lewis, and X.~Wang, ``Deep learning for
  real-time atari game play using offline monte-carlo tree search planning,''
  in \emph{Advances in Neural Information Processing Systems}.\hskip 1em plus
  0.5em minus 0.4em\relax Curran Associates, Inc., 2014, pp. 3338--3346.

\bibitem{IEEEexample:mnih2015human}
{V. Mnih \textit{et al.}}, ``Human-level control through deep reinforcement
  learning,'' \emph{Nature}, vol. 518, no. 7540, pp. 529--533, 2015.

\bibitem{IEEEexample:simonyan2014very}
K.~Simonyan and A.~Zisserman, ``Very deep convolutional networks for
  large-scale image recognition,'' \emph{arXiv preprint arXiv:1409.1556}, 2014.

\bibitem{IEEEexample:levelling}
M.~Zaffar, A.~Khaliq, S.~Ehsan, M.~Milford, and K.~D. McDonald-Maier,
  ``Levelling the playing field: A comprehensive comparison of visual place
  recognition approaches under changing conditions,'' \emph{arXiv preprint
  arXiv:1903.09107}, 2019.

\bibitem{IEEEexample:lstm}
S.~Hochreiter and J.~Schmidhuber, ``Long short-term memory,'' \emph{Neural
  Computation}, vol.~9, pp. 1735--1780, 1997.

\bibitem{IEEEexample:ppo}
J.~Schulman, F.~Wolski, P.~Dhariwal, A.~Radford, and O.~Klimov, ``Proximal
  policy optimization algorithms,'' \emph{arXiv preprint arXiv:1707.06347},
  2017.

\bibitem{IEEEexample:curriculumlearning}
Y.~Bengio, J.~Louradour, R.~Collobert, and J.~Weston, ``Curriculum learning,''
  in \emph{Proceedings of the 26th Annual International Conference on Machine
  Learning}, 2009, pp. 41--48.

\bibitem{IEEEexample:impala}
{L. Espeholt \textit{et al.}}, ``{IMPALA: Scalable Distributed Deep-RL with
  Importance Weighted Actor-Learner Architectures},'' \emph{arXiv preprint
  arXiv:1802.01561}, 2018.

\bibitem{IEEEexample:cueva2018emergence}
C.~J. Cueva and X.-X. Wei, ``Emergence of grid-like representations by training
  recurrent neural networks to perform spatial localization,'' \emph{arXiv
  preprint arXiv:1803.07770}, 2018.

\end{thebibliography}

\end{document}